\documentclass[journal]{IEEEtran}
\usepackage{cite}

\ifCLASSINFOpdf
 \usepackage[pdftex]{graphicx, xcolor}
\else
\fi

\usepackage{amsmath}

\usepackage{algorithm,algorithmic}

\usepackage{array}

\ifCLASSOPTIONcompsoc
 \usepackage[caption=false,font=normalsize,labelfont=sf,textfont=sf]{subfig}
\else
 \usepackage[caption=false,font=footnotesize]{subfig}
\fi

\usepackage{url}

\graphicspath{{figs/}}
\usepackage{siunitx}
\usepackage{enumitem}
\usepackage{booktabs}
\usepackage{etoolbox}
\usepackage{multirow}
\usepackage[export]{adjustbox}

\usepackage{bbm} 

\newcommand{\Review}[1]{\textcolor{black}{#1}}
\newcommand{\RA}[1]{\textcolor{black}{#1}}
\newcommand{\RB}[1]{\textcolor{black}{#1}}
\newcommand{\RC}[1]{\textcolor{black}{#1}}
\newcommand{\RD}[1]{\textcolor{black}{#1}}

\begin{document}

\title{Block-wise Image Transformation with Secret Key for Adversarially Robust Defense}

\author{{MaungMaung~AprilPyone, Hitoshi~Kiya}
 \thanks{The authors are with the Department of Computer Science, Tokyo Metropolitan University, Tokyo 191-0065, Japan (email: april-pyone-maung-maung@ed.tmu.ac.jp, kiya@tmu.ac.jp).}
}

\maketitle

\begin{abstract}
In this paper, we propose a novel defensive transformation that enables us to maintain a high classification accuracy under the use of both clean images and adversarial examples for adversarially robust defense. The proposed transformation is a block-wise preprocessing technique with a secret key to input images. We developed three algorithms to realize the proposed transformation: Pixel Shuffling, Bit Flipping, and FFX Encryption. Experiments were carried out on the CIFAR-10 and ImageNet datasets by using both black-box and white-box attacks with various metrics including adaptive ones. The results show that the proposed defense achieves high accuracy close to that of using clean images even under adaptive attacks for the first time. In the best-case scenario, a model trained by using images transformed by FFX Encryption (block size of 4) yielded an accuracy of \SI{92.30}{\percent} on clean images and \SI{91.48}{\percent} under PGD attack with a noise distance of 8/255, which is close to the non-robust accuracy (\SI{95.45}{\percent}) for the CIFAR-10 dataset, and it yielded an accuracy of \SI{72.18}{\percent} on clean images and \SI{71.43}{\percent} under the same attack, which is also close to the standard accuracy (\SI{73.70}\percent) for the ImageNet dataset. Overall, all three proposed algorithms are demonstrated to outperform state-of-the-art defenses including adversarial training whether or not a model is under attack.
\end{abstract}

\begin{IEEEkeywords}
Adversarial Defense, Image Encryption, Image Classification.
\end{IEEEkeywords}

\IEEEpeerreviewmaketitle

\section{Introduction}
\begin{figure}[!t]
\centering
\includegraphics{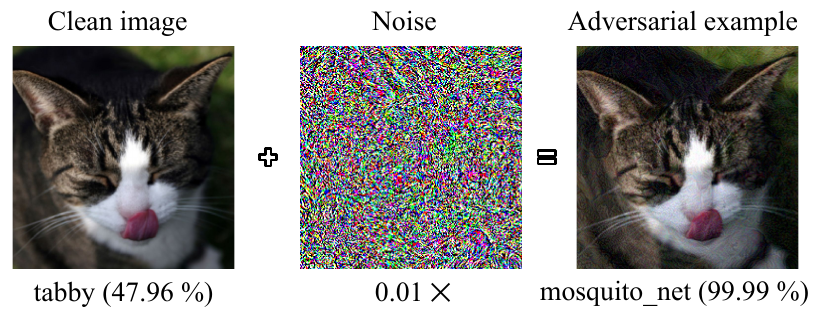}
\caption{Adversarial example.\label{fig:adv-ex}}
\end{figure}

Although deep neural networks (DNNs) have lead to major breakthroughs in computer vision, for a wide range of applications, where safety and security are critical, there is concern about their reliability. DNNs in general suffer from attacks such as model inversion attacks~\cite{fredrikson2015model}, membership inference attacks~\cite{shokri2017membership}, and adversarial attacks~\cite{Szegedy14}. In particular, carefully perturbed data points known as adversarial examples are indistinguishable from clean data points, but they cause DNNs to make erroneous predictions~\cite{Szegedy14,Biggio13}. As an example, in Fig.~\ref{fig:adv-ex}, the network here classified the clean image correctly as ``tabby'' with a \SI{47.96}{\percent} probability. After adding a small fraction of noise, the network misclassified the tabby cat as ``mosquito\_net'' with \SI{99.99}{\percent} confidence. Adversarial examples create a rising concern where DNNs are to be deployed in security-critical applications such as autonomous vehicles, speech recognition, natural language processing, and malware detection. Therefore, a lot of effort has been put towards adversarial robustness.

Researchers have proposed numerous ways of constructing adversarial examples. Such works include~\cite{Szegedy14,Goodfellow15,Kurakin2017,Dezfooli16,Carlini017,Madry18}, in which $\ell_{p}$-bounded perturbation has been found. In the context of computer vision, these threat models do not match real world applications~\cite{engstrom2017rotation,gilmer2018motivating} because there can be various physical conditions (e.g., camera angle, lighting/weather), physical limits on imperceptibility, etc. However, it has been proved that adversarial threats on neural networks remain real~\cite{Eykholt18,Athalye18,KurakinGB17a,Papernot17,Sharif2016}. In addition, $\ell_{p}$-bounded threat models are crucial for principled deep learning due to their well-defined nature~\cite{Carlini19}. They are helpful not only for evaluating the robustness of deep learning models but also for understanding them better. It is almost certain that models that are not robust against $\ell_{p}$-bounded attacks will fail in real world scenarios.

With the development of adversarial attacks, numerous adversarial defenses have been proposed in the literature. To the best of our knowledge, there is no robust model that has a similar accuracy to a non-robust one. Some of the most reliable defenses are certified ones and adversarial training. However, certified defenses are not scalable, and the accuracy of adversarial training is not comparable to that of standard training. Alternatively, researchers have also come up with preprocessing approaches to improve the classification accuracy. Unfortunately, most of these approaches are broken by powerful adaptive attacks~\cite{AthalyeC018}. Therefore, finding ways to achieve high accuracy and adversarial robustness is a growing concern and an on-going area of research with a high demand for computer vision because of the wide range of applications.

\RA{More importantly, adversarial attacks and defenses have entered into an arms race in the literature. New defenses are also broken by performing adaptive attacks~\cite{Trammer-Arxiv-2020}. Conventional adversarial defenses either reduce classification accuracy significantly or are completely broken. Therefore, in this work, we aim to achieve a high classification accuracy not only for clean examples but also for adversarial ones.}

We propose a block-wise defensive transformation with a secret key that is inspired by perceptual image encryption techniques such as~\cite{Chuman-TIFS-2019,Warit-APSIPAT-2019,2019-ICIP-Warit,Warit-Access-2019,madono2020block,Tanaka-ICCETW-2018,Kurihara-IEICE-2017}. \RA{ Modern deep convolutional neural networks such as ResNet are known to be sensitive to small image transformation~\cite{2019-JMLR-Azulay}. Therefore, many researchers have been seeking learnable image encryption methods~\cite{2019-ICIP-Warit,Warit-Access-2019,madono2020block,Tanaka-ICCETW-2018} that do not cause big drops in accuracy for privacy-preserving DNNs, but they have not considered robustness against adversarial examples.} Deriving from such encryption methods, we develop three block-wise transformation algorithms to carry out the proposed transformation: \RC{Pixel Shuffling, Bit Flipping, and FFX Encryption}. The proposed transformation is utilized to transform training/test images as a preprocessing technique, and a model is trained/tested by the transformed images. In addition, we also design adaptive attacks while accounting for obfuscated gradients~\cite{AthalyeC018} to evaluate models trained by the proposed transformation algorithms. As a result, the models trained by the proposed transformation make correct predictions for both clean images and adversarial examples. We make the following contributions in this paper.
\begin{itemize}
\item We apply extended perceptual image encryption techniques with a secret key to adversarial defenses for the first time.
\item We develop three block-wise transformation algorithms: \RC{Pixel Shuffling, Bit Flipping, and FFX Encryption.}
\item We conduct extensive experiments on both black-box and white-box attacks including adaptive ones and present empirical results to show the effectiveness of the proposed defense.
\end{itemize}
In experiments, the proposed defense is confirmed to outperform state-of-the-art adversarial defenses. A part of this work (Pixel Shuffling) was introduced in~\cite{Maung-ICIP-2020}. We not only evaluate Pixel Shuffling under both black-box and white-box attacks with different metrics, and adaptive attacks with key estimation approaches but also introduce other novel algorithms in this paper.

The rest of this paper is structured as follows. Section~\ref{sec:related-work} presents related work on adversarial attacks and defenses.
Section~\ref{sec:threat} describes threat models.
Regarding the proposed defense, Section~\ref{sec:proposed} includes notations, an overview, the three proposed block-wise transformation algorithms, the properties of block-wise transformations with keys, and a discussion on key management and robustness against adaptive attacks. Experiments on various attacks including adaptive ones are presented in Section~\ref{sec:experiments}, and Section~\ref{sec:conclusion} concludes this paper.

\section{Related Work\label{sec:related-work}}
\subsection{Adversarial Attacks}
The goals of adversarial attacks on neural networks are confidence reduction, misclassification, and targeted misclassification. The attacks can be divided into two categories: poisoning/causative attacks (i.e., training time attacks) and evasion/exploratory attacks (i.e., test time attacks)~\cite{barreno2010security}. Poisoning attacks happen during training time, where an adversary introduces crafted malicious examples into training data to manipulate the behavior of models. Even one single poisonous image can compromise a model when transfer learning is used~\cite{shafahi2018poison}. Evasion attacks are also called ``adversarial examples,'' in which crafted imperceptible perturbations are added. In this work, we focus on defending against evasion attacks. 

Traditionally, evasion attacks are classified into three groups based on the knowledge of a particular model and training data available to the adversary: white-box, black-box, and gray-box. Under white-box settings, the adversary has direct access to the model, its parameters, training data, and defense mechanism. However, the adversary does not have any knowledge on the model, except the output of the model in black-box attacks. Between white-box and black-box methods, there are gray-box attacks that imply that the adversary knows something about the system (i.e., partial knowledge of the model such as its architecture, parameters, or training data).

Under white-box settings, given an input image $x$ and a classifier $f_\theta(\cdot)$ parameterized by $\theta$, an adversarial example $x'$ is constructed such that $f_\theta(x') \neq y$, where $y$ is a true class. This is done by minimizing the perturbation $\delta$,
\begin{equation}
 \underset{\delta}{\text{minimize}} \left\rVert \delta \right\rVert_p, \;\;\text{s.t.}\;\; f_\theta(x + \delta) \neq y,
\end{equation}
or by maximizing the loss function,
\begin{equation}
 \underset{\delta \in \Delta}{\text{maximize}}\; \mathcal{L}(f_\theta(x + \delta), y).
\end{equation}
Usually, a typical threat model is bounded by an $\ell_p$ norm such that $\Delta = \{\delta: \left\rVert \delta \right\rVert_p \leq \epsilon \}$ for some perturbation distance $\epsilon > 0$.

One of the easy and popular ways of generating adversarial examples is the fast gradient sign method (FGSM)~\cite{Goodfellow15} under an $\ell_\infty$ norm with a single gradient step. Its iterative version is the basic iterative method (BIM)~\cite{Kurakin2017}. BIM with multiple random restarts and initialization with uniform random noise is recognized as a projected gradient descent (PGD)~\cite{Madry18} adversary. There are other iterative optimization-based attacks such as the Carlini and Wagner attack (CW)~\cite{Carlini017} for the $\ell_2$ bounded metric and the elastic-net attack (EAD)~\cite{Chen-AAAI-2018} for the $\ell_1$ bounded metric. CW finds the smallest noise under the $\ell_2$ metric with a new loss function. CW is also a special case of EAD, where the $\ell_1$ regularization parameter is set to zero~\cite{Chen-AAAI-2018}. In this work, we utilize three state-of-the-art attacks (PGD, CW, and EAD) to generate different sets of adversarial examples under $\ell_\infty$, $\ell_2$, and $\ell_1$ metrics to evaluate the proposed defense.

\RA{Under black-box settings, the above white-box attacks can be applied via a substitute model. Several techniques have been proposed to improve transferability with this type of black-box attack~\cite{dong2018boosting,dong2019evading,xie2019improving}. Moreover, there are also gradient-free methods that estimate gradients such as~\cite{chen2017zoo,ilyas2018black,uesato2018adversarial,cheng2019improving}. Another recent black-box attack, NATTACK, learns a probability distribution centered around the input such that a sample drawn from that distribution is likely an adversarial example~\cite{pmlr-v97-li19g}. Additionally, the OnePixel attack constructs an adversarial example by modifying one or a few pixels without accessing the weights of the model with differential evolution~\cite{su2019one}. In this work, we employ the OnePixel attack~\cite{su2019one}, NATTACK~\cite{pmlr-v97-li19g}, and one of the gradient estimation attacks, SPSA~\cite{uesato2018adversarial}, to evaluate the proposed defense under black-box settings.}

\subsection{Adversarial Defenses}
\begin{figure}[!t]
\centering
\includegraphics{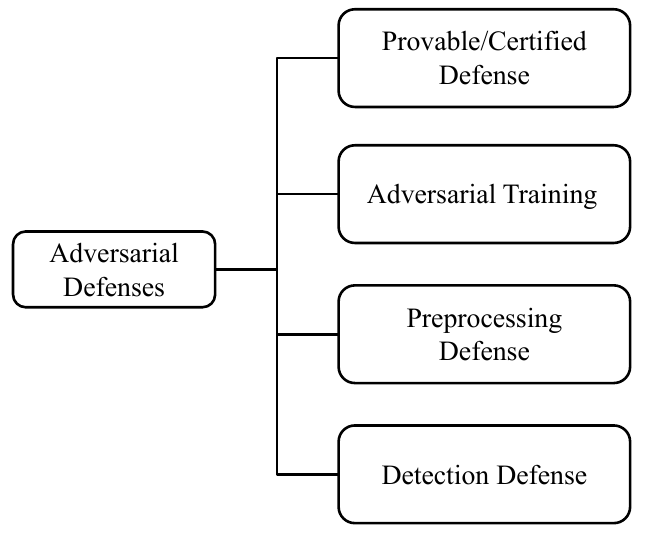}
\caption{Different approaches for adversarial defense.\label{fig:tifs-defenses}}
\end{figure}

The goal of a defense method is to make a model that is accurate not only for clean input but also for adversarial examples. There are many different approaches to achieving this goal, such as certified and provable defenses, adversarial training, preprocessing techniques, and detection algorithms, as shown in Fig.~\ref{fig:tifs-defenses}.

Ideally, provable defenses are desired. Inspiring works such as~\cite{Raghunathan18,Dvijotham18,Wong17} proposed provable secure training. Although these methods are attractive, they are not scalable. Some certified defenses have been scaled to a certain degree~\cite{Salman-NIPS-2019, Gowal-Arxiv-2018, Mirman-ICML-2018, Wong18}, but the accuracy is still not comparable to empirically robust models.

Current state-of-the-art empirically robust defenses are under the use of adversarial training. The earliest form of adversarial training is to inject FGSM-based adversarial noise into the training data~\cite{Goodfellow15}. Since FGSM is not iterative and not robust against iterative attacks such as PGD, FGSM-based training was found to be ineffective~\cite{Kurakin2017,Madry18}. Madry et al.\ proposed adversarial training with a PGD adversary, achieving the best empirical robustness to date~\cite{Madry18}. However, PGD training is computationally expensive. To make the computation of adversarial training more feasible, ``free'' adversarial training was proposed in which gradients are computed with respect to the network parameters and the input image on the same backward pass~\cite{Shafahi-NIPS-2019}. In addition to ``free'' adversarial training, ``fast'' adversarial training was proposed and uses FGSM and standard efficient training tricks~\cite{Wong-ICLR-2020}. Although FGSM-based adversarial training was dismissed before, it is shown to be effective when random initialization is introduced~\cite{Wong-ICLR-2020}. Nevertheless, while adversarial training is repeatedly found to be robust against the best known adversaries~\cite{AthalyeC018}, the accuracy is still very low compared with non-robust models.

Another approach to adversarial defenses is the use of preprocessing techniques. The works that take this direction utilize various ways of transformation such as thermometer encoding~\cite{Buckman18}, image processing-based techniques~\cite{Guo17, Xie17}, making small changes to pixels with the intent of removing adversarial noise~\cite{Song17}, and GAN-based transformation~\cite{Samangouei18}. These preprocessing defenses are appealing at first due to their higher accuracy. However, they have all been broken because these conventional preprocessing defenses rely on obfuscated gradients by~\cite{AthalyeC018}. Accounting for this problem, Raff et al.\ came up with a preprocessing defense that uses a number of random different transforms with random parameters~\cite{Raff19}. Although their work claims majorly improved accuracy on ImageNet, applying many transforms for each image is computationally expensive and reduces the accuracy when the model is not under attack. In addition, one work enforces the use of 1-bit dithered images for training and testing a model~\cite{Maung-Access-2019}. However, typical cameras capture an image in 8-bit, and the use of 1-bit limits the range of application scenarios.

Moreover, instead of defending against adversarial examples directly, there are defenses to detect adversarial examples. Metzen et al.\ proposed a detection method that trains a binary classification network to distinguish clean data from adversarial examples~\cite{Metzen-ICLR-2017}. Another work by~\cite{Feinman-Arxiv-2017} detects adversarial examples by looking at the features in the subspace of deep neural networks. However, it is reported that detection methods can also be bypassed~\cite{Carlini-AISec-2017}.

All in all, adversarial defense approaches decrease the classification accuracy of a model. Even worse, most of the defenses, especially preprocessing-based methods, are defeated due to obfuscated gradients~\cite{AthalyeC018} \RB{and do not embed a secret key into the model inference process.}
Therefore, attaining robust as well as high accuracy remains an open problem in adversarial defense research.

In this work, we approach adversarial defense in a different way by taking inspiration from perceptual image encryption techniques such as~\cite{Chuman-TIFS-2019,Warit-APSIPAT-2019,Warit-Access-2019,madono2020block,Tanaka-ICCETW-2018}. Perceptual image encryption techniques have never been applied before in this line of work. Similar to our work (Pixel Shuffling), Taran et al.\ first introduced a pixel shuffling approach (pixel-wise manner) with a secret key by using a standard random permutation~\cite{Taran-ECCV-2018}.
\RB{Although their method~\cite{Taran-ECCV-2018} was effective to defend against adversarial examples, it was tested only on small datasets (MNIST~\cite{LeCun-IEEE-1998} and F-MNIST~\cite{Xiao-preprint-2017}) and clean accuracy is significantly dropped on larger datasets such as CIFAR-10\cite{Krizhevsky09} and ImageNet~\cite{ILSVRC15}. The reason is that shuffling in a pixel-wise manner loses spatial perceptual information.
In contrast, the proposed algorithm (Pixel Shuffling) is block-wise pixel shuffling and designed to maintain a high clean accuracy.}
In this paper, we will show that the extension of perceptual image encryption techniques is effective in defending against adversarial examples.

\section{Threat Models\label{sec:threat}}
\RB{The goal of an adversarial defense is to keep the classification accuracy on both clean images and adversarial examples high. To evaluate a defense method, precisely defining threat models is necessary. A threat model includes a set of assumptions such as an adversary's goals, knowledge, and capabilities~\cite{Carlini19}. We also define attack scenarios considering practical applications.}

\subsection{Adversary's Goals}
\RB{An adversary can construct adversarial examples to achieve different goals when attacking a model: whether to reduce the performance accuracy (i.e., untargeted attacks) or to classify a targeted class (i.e., targeted attacks). Formally, untargeted attacks will cause a classifier $f_\theta$ to misclassify a true class $y_{\text{true}}$, given an adversarial example $x'$ (i.e., $f_{\theta}(x') \neq y_{\text{true}}$), and targeted ones will force the classifier to a targeted label (i.e., $f_{\theta}(x') = y_{\text{targeted}}$). In this paper, we focus on untargeted attacks, although targeted attacks can be launched in a similar fashion.}

\subsection{Adversary's Knowledge}
\RB{According to~\cite{Carlini19}, the adversary's knowledge can be white-box (inner workings of the defense mechanism, complete knowledge on the model and its parameters), black-box (no knowledge on the model) and gray-box, that is, anything in between white-box and black-box. As in the field of cryptography, there can be a small amount of secret information even in white-box settings if the secret information must be easily replaceable and non-extractable~\cite{Carlini19}. For our proposed defense, we introduce a secret key with a transparent algorithm for the first time. The secret key can be replaced by retraining the model, and it cannot be extracted from the training data nor the model. The key is utilized to preprocess input on the fly just before the input goes into the model. In this work, we consider both white-box and black-box attacks while keeping a secret key.}

\subsection{Adversary's Capabilities}
\RB{Depending on the requirements of different applications, a secret key may or may not be required for inference. However, it should be securely stored or distributed. We assume the adversary does not have access to information with respect to the secret key (either the key itself or model output with respect to the correct secret key). However, the adversary may guess/estimate the secret key and observe the model. Then, they can perform untargeted attacks in which small changes are made under different metrics ($\ell_{0}$, $\ell_{1}$, $\ell_{2}$, $\ell_{\infty}$) that change the true class of the input.}

\subsection{Attack Scenarios}
\RB{We consider the following practical application scenarios.}

\RB{\textbf{Black-box:} The attacker queries the protected model with their key and observes the output of the model. Specifically, the attacker performs three powerful black-box attacks: OnePixel~\cite{pmlr-v97-li19g}, NATTACK~\cite{su2019one}, and SPSA~\cite{uesato2018adversarial}.}

\RB{\textbf{White-box:} In the proposed defense, the key is not a part of the model parameters. The model may be stolen in the case of sharing the model. We assume a scenario in which the model weights and the defense algorithm are available to the attacker. Since the defense algorithm is known, the attacker may carry out white-box attacks with their key. Specifically, the attacker carries out three strong white-box attacks: PGD~\cite{Madry18}, CW~\cite{Carlini017}, and EAD~\cite{Chen-AAAI-2018}. To make the attacks more successful, we assume the attacker incorporates the defense algorithm with an unknown key during the attacks. In other words, the white-box attacks are run on top of the defense algorithm with the attacker's key.}

\section{Proposed Defense\label{sec:proposed}}
Image classification is the task of classifying an input image into a class category according to its visual content. The proposed defense targets robust predictions in the image classification task, which is a core problem in computer vision.

\subsection{Notation}
The following notations are utilized throughout this paper.
\begin{itemize}
\item $w$, $h$, and $c$ are used to denote the width, height, and the number of channels of an image.
\item The tensor $x \in {[0, 1]}^{c \times w \times h}$ represents an input color image.
\item $\delta$ denotes adversarial noise.
\item $\delta_{a}$ denotes adaptive adversarial noise.
\item The tensor $x_t \in {[0, 1]}^{c \times w \times h}$ represents a transformed image.
\item $M$ is the block size of an image.
\item Tensors $x_b, x_b' \in {[0, 1]}^{w_b \times h_b \times p_b}$ are a block image and a transformed block image, respectively, where $w_b = \frac{w}{M}$ is the number of blocks across width $w$, $h_b = \frac{h_b}{M}$ is the number of blocks across height $h$, and $p_b = M \times M \times c$ is the number of pixels in a block.
\item A pixel value in a block image ($x_b$ or $x_b'$) is denoted by $x_b(i, j, k)$ or $x_b'(i, j, k)$, where $i \in \{0, \dots, w_b - 1\}$, $j \in \{0, \dots, h_b - 1\}$, and $k \in \{0, \dots, p_b - 1\}$ are indices corresponding to the dimension of $x_b$ or $x_b'$.
\item $B$ is a block of an image, and its dimension is $M \times M \times c$.
\item $\hat{B}$ is a flattened version of block $B$, and its dimension is $1 \times 1 \times p_b$.
\item An encryption key is denoted by $K$.
\item A password required for format-preserving encryption, which refers to encrypting in such a way that the output is in the same format as its input, is denoted as $P$.
\item $\text{Enc}(n, P)$ denotes format-preserving Feistel-based encryption (FFX)~\cite{Bellare-NIST-2010} with a length of $3$, where $n$ is an integer (used only in FFX Encryption).
\item A classifier with parameters $\theta$ is denoted as $f_{\theta}(\cdot)$.
\end{itemize}

\subsection{Overview}
\RA{We propose a general key-based adversarial defense that satisfies two requirements: defending against adversarial examples and maintaining a high classification accuracy. Assuming the key stays secret, an attacker will not obtain any useful information on the model, which will render the adversarial attack ineffective. The main idea of the proposed method is to embed a secret key into the model structure with minimal impact on model performance. To maintain a high classification accuracy, the proposed defense is designed in such a way that each block position in an input image is not changed.}

\begin{figure*}[!t]
\centering
\subfloat[]{\includegraphics[width=0.45\linewidth]{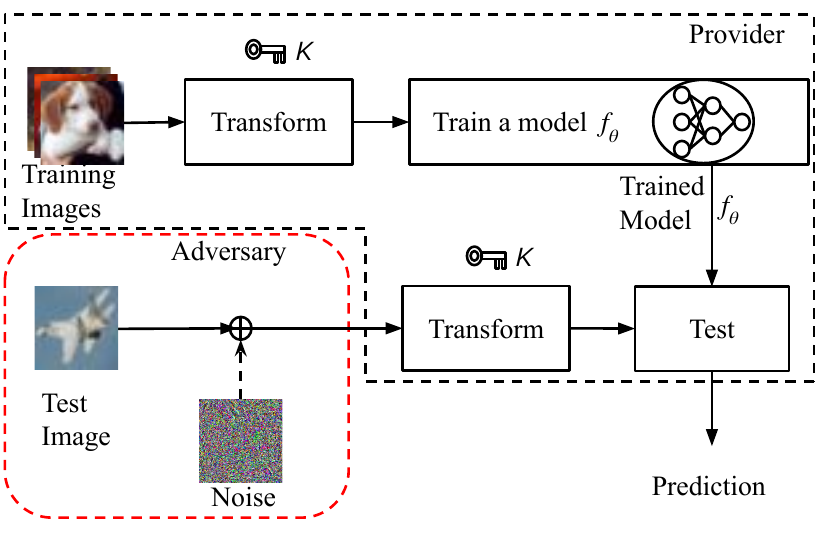}%
\label{fig:scenario-a}}
\hfil
\subfloat[]{\includegraphics[width=0.45\linewidth]{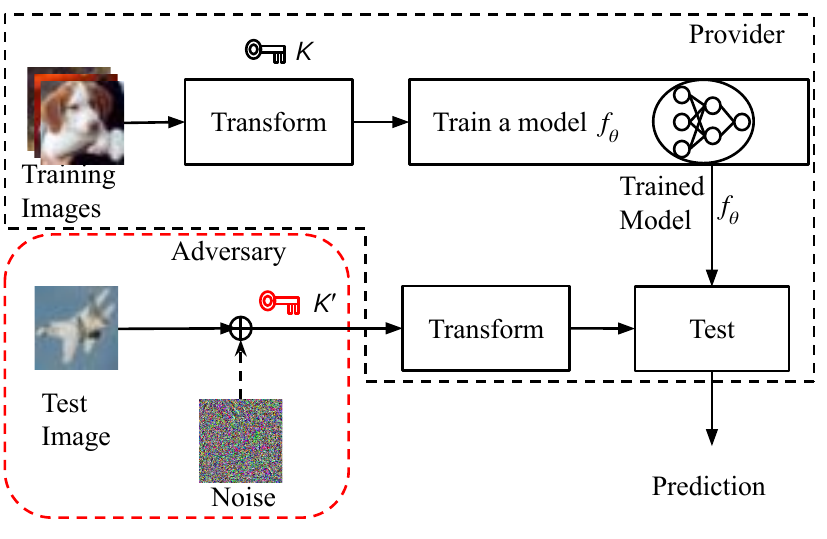}%
\label{fig:scenario-b}}
\caption{Scenarios of image classification with proposed defense. (a) Scenario A where key $K$ is saved with provider. (b) Scenario B where key $K$ is required from user/adversary by provider for inference.\label{fig:overview}}
\end{figure*}

\RB{Based on different types of key management, the proposed defense can be applied in two scenarios as shown in Fig.~\ref{fig:overview}: (1) Scenario A, where key $K$ is saved with a provider, and (2) Scenario B, where key $K$ is required by a provider for inference. As an example, Scenario A can be deployed in self-driving cars, and Scenario B can be utilized in vision application programming interfaces (APIs).}

\begin{figure*}[!t]
\centering
\includegraphics{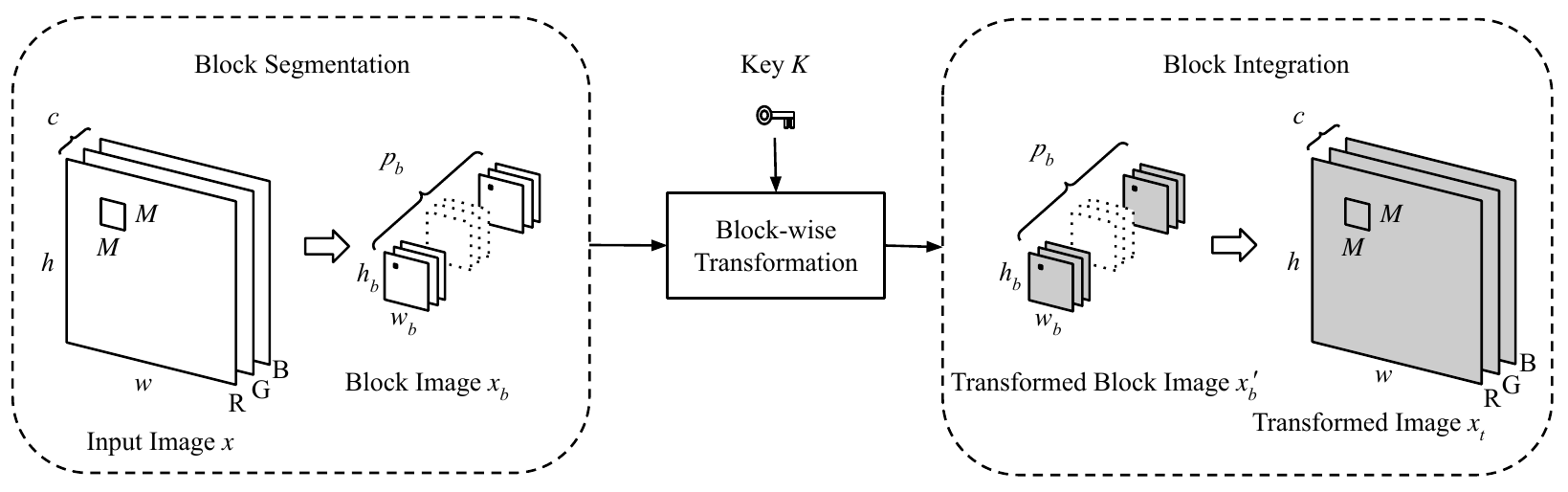}
\caption{Process of proposed transformation.\label{fig:tifs-proposed}}
\end{figure*}

\begin{figure*}[!t]
\centering
\includegraphics{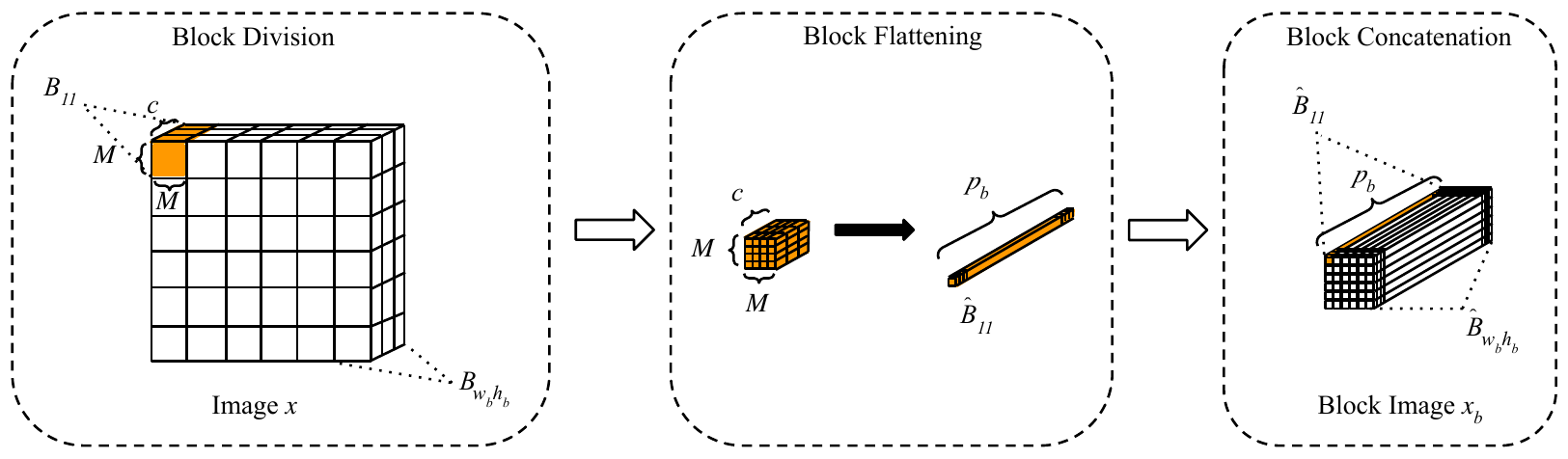}
\caption{Process of block segmentation.\label{fig:tifs-segmentation}}
\end{figure*}

The proposed defense is a preprocessing technique that transforms an input image with a secret key in a block-wise manner. Both training and testing images are transformed with a secret key prior to training or testing by the provider. Generally, there are three parts to the proposed defensive transformation: block segmentation, block-wise transformation, and block integration (see Fig.~\ref{fig:tifs-proposed}). The process of the proposed transformation is shown as follows.
\begin{enumerate}
\item \textbf{Block Segmentation:} The process of block segmentation is illustrated in Fig.~\ref{fig:tifs-segmentation}.
\begin{itemize}
\item An input image $x$ is divided into blocks such that $\{B_{11}, B_{12}, \dots, B_{w_{b}h_{b}}\}$.
\item Each block in $x$ is flattened to obtain $\{\hat{B}_{11}, \hat{B}_{12}, \dots, \hat{B}_{w_{b}h_{b}}\}$.
\item The flattened blocks are concatenated in such a way that the relative spatial location among blocks in $x_b$ is the same as that among blocks in $x$.
\end{itemize}
\item \textbf{Block-wise Transformation} Given a secret key $K$ that is a seed for generating a pseudo random integer vector with a size of $p_b$, $x_b$ is transformed by using a block-wise transformation algorithm, $t(x_b, K)$. The transformed block image is written as
\begin{equation}
 x_b' = t(x_b, K). \label{eq:proposed}
\end{equation}
\item \textbf{Block Integration} The transformed blocks in $x_b'$ are integrated back to the original dimension (i.e., $c \times w \times h$) in the reverse order to the block segmentation process for obtaining a transformed image $x_t$.
\end{enumerate}

\subsection{Three Proposed Block-wise Transformations}
We propose three block-wise transformation algorithms, \RC{Pixel Shuffling, Bit Flipping, and FFX Encryption}, for realizing $t(x_b, K)$. A block-wise transformation takes a block image $x_b$ and a key $K$ and then outputs a transformed block image $x_b'$. In the case of FFX Encryption, there is an additional parameter, password $P$, for format-preserving encryption. Although one of the three algorithms (Pixel Shuffling) was discussed in our previous work~\cite{Maung-ICIP-2020}, we extend the evaluation of Pixel Shuffling with various white-box and black-box attacks under different metrics and adaptive attacks with key estimation in this paper.

\textbf{Pixel Shuffling:} There are two steps to pixel shuffling as described in Algorithm~\ref{algo:shuffle}:
\begin{enumerate}
\item Generate a random permutation vector $v = (v_0, v_1, \dots, v_k, \dots, v_{k'}, \dots, v_{p_b - 1})$ that consists of randomly permuted integers from $0$ to $p_b - 1$ by using key $K$. Let $k, k' \in \{0, \dots, p_b - 1\}$ and $v_k \neq v_{k'}$ if $k \neq k'$.
\item Perform block-wise shuffling on the basis of $v$. Basically, positions of pixel values in each block are changed on the basis of $v$, i.e.,
 \begin{equation}
 x_b'(i, j, v_k) = x_b(i, j, k).
 \end{equation}
\end{enumerate}

\begin{algorithm}
\caption{Pixel Shuffling\label{algo:shuffle}}
 \begin{algorithmic}
 \renewcommand{\algorithmicrequire}{\textbf{Input:}}
 \renewcommand{\algorithmicensure}{\textbf{Output:}}
 \REQUIRE{$x_b, K$}
 \ENSURE{$x_b'$}
 \STATE{Generate a random permutation vector $v$ by $K$}
 \STATE{$x_b' \leftarrow x_b[:, :, v]$}
 \end{algorithmic}
\end{algorithm}

\textbf{Bit Flipping:} There are four steps to pixel intensity inversion as described in Algorithm~\ref{algo:inverse}:
\begin{enumerate}
\item Generate a random binary vector $r = (r_0, r_1, \dots, r_k, \dots, r_{p_b -1})$, $r_k \in \{0, 1\}$ by using key $K$. To keep the transformation consistent, $r$ is distributed with \SI{50}{\percent} of ``0''s and \SI{50}{\percent} of ``1''s.
\item Convert every pixel value to be in $255$ scale with 8 bits (i.e., multiply $x_b$ by $255$).
\item Perform block-wise negative/positive transformation on the basis of $r$. Basically, every pixel value in block $\hat{B}_{ij}$ is applied to
 \begin{equation}
 x_b'(i, j, k) = \left\{
 \begin{array}{ll}
 x_b(i, j, k) & (r_k = 0)\\
 x_b(i, j, k) \oplus (2^L - 1) & (r_k = 1),
 \end{array}
 \right.
 \end{equation}
where $L$ is the number of bits used in $x_b(i, j, k)$, and $L = 8$ is used in this paper.
\item Convert every pixel value back to $[0, 1]$ scale (i.e., divide $x'_b$ by $255$).
\end{enumerate}

\begin{algorithm}
\caption{Bit Flipping\label{algo:inverse}}
 \begin{algorithmic}
 \renewcommand{\algorithmicrequire}{\textbf{Input:}}
 \renewcommand{\algorithmicensure}{\textbf{Output:}}
 \REQUIRE{$x_b$, $K$}
 \ENSURE{$x_b'$}
 \STATE{Generate a random binary vector $r$ by $K$}
 \STATE{// \textit{Make pixel values be at 255 scale}}
 \STATE{$x_b \leftarrow x_b \cdot 255$}
 \STATE{$x_b'[:, :, r] \leftarrow 255 - x_b[:, :, r]$}
 \STATE{$x_b' \leftarrow x_b' / 255$}
 \end{algorithmic}
\end{algorithm}

\textbf{FFX Encryption:} \RB{In Bit Flipping, there are only two possibilities: whether the intensity of a pixel value is inversed or not. In contrast, we replace Bit Flipping with a cryptographic property (i.e., FFX mode) to generate a unique pattern in a block-wise manner, where the number of patterns is much larger than that of Bit Flipping. For this reason, FFX Encryption is applied to adversarial defense for the first time.}

Apart from key $K$, FFX-based transformation also requires a password $P$ for format-preserving Feistel-based encryption (FFX)~\cite{Bellare-NIST-2010}. The pixel value $x_b(i, j, k) \in \{0, 1, \dots, 254, 255\}$ is encrypted by FFX with a length of $3$ digits to cover the whole range from $0$ to $255$. \RB{FFX randomly transforms each pixel with an integer value of ($0$--$255$) into a pixel with an integer value of ($0$--$999$) preserving the integer format.} There are four steps to FFX-based transformation as described in Algorithm~\ref{algo:encrypt}:
\begin{enumerate}
\item Generate a random binary vector $r = (r_0, r_1, \dots, r_k, \dots, r_{p_b -1})$, $r_k \in \{0, 1\}$ by using key $K$. To keep the transformation consistent, $r$ is distributed with \SI{50}{\percent} of ``0''s and \SI{50}{\percent} of ``1''s.
\item Convert every pixel value to be at $255$ scale with 8 bits (i.e., multiply $x_b$ by $255$).
\item Perform block-wise FFX-based transformation on the basis of $r$ and $P$. Basically, every pixel value in block $\hat{B}_{ij}$ is applied to
 \begin{equation}
 x_b'(i, j, k) = \left\{
 \begin{array}{ll}
 x_b(i, j, k) & (r_k = 0)\\
 \text{Enc}(x_b(i, j, k), P) & (r_k = 1).
 \end{array}
 \right.
 \end{equation}
\item Convert every pixel value back to $[0, 1]$ scale (i.e., divide $x'_b$ by the maximum value of $x'_b$).
\end{enumerate}

\RB{On a side note, only pixel values of 0 to 255 are encrypted once, and the block-wise transformation uses a lookup table. Therefore, the computational cost of encrypting 256 integer values in FFX mode is negligible and does not cause any significant overheads when training or testing a model.}

\begin{algorithm}
\caption{FFX Encryption\label{algo:encrypt}}
 \begin{algorithmic}
 \renewcommand{\algorithmicrequire}{\textbf{Input:}}
 \renewcommand{\algorithmicensure}{\textbf{Output:}}
 \REQUIRE{$x_b$, $K$, $P$}
 \ENSURE{$x_b'$}
 \STATE{Generate a random binary vector $r$ by $K$}
 \STATE{// \textit{Make pixel values be at 255 scale}}
 \STATE{$x_b \leftarrow x_b \cdot 255$}
 \STATE{$x_b'[:, :, r] \leftarrow \text{Enc}(x_b[:, :, r], P$)}
 \STATE{max $\leftarrow$ the maximum value of the encryption}
 \STATE{$x_b' \leftarrow x_b' / \text{max}$}
 \end{algorithmic}
\end{algorithm}

\subsection{Properties of Block-wise Transformation with Key\label{sec:difference}}
\begin{figure}[!t]
\centering
\subfloat[]{\includegraphics[width=0.25\linewidth]{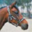}%
\label{fig:horse}}
\hfil
\subfloat[]{\includegraphics[width=0.25\linewidth]{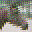}%
\label{fig:horse-s}}
\hfil
\subfloat[]{\includegraphics[width=0.25\linewidth]{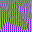}%
\label{fig:horse-i}}
\hfil
\subfloat[]{\includegraphics[width=0.25\linewidth]{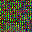}%
\label{fig:horse-e}}
\caption{Example of images generated by three proposed transformations with $M = 2$. (a) Original image. (b) Pixel Shuffling. (c) Bit Flipping. (d) FFX Encryption.\label{fig:vis-diff}}
\end{figure}

A classifier model, $f_{\theta}(\cdot)$, trained by using transformed images is affected by both key $K$ and the block-wise transformation used for transforming images. Each transformation algorithm creates a unique pattern as illustrated in Fig.~\ref{fig:vis-diff}, where a test image (``horse'') was transformed by the three proposed algorithms. The pattern created by the defensive transformation makes the gradients of the loss function with respect to the parameters unique to the particular transformation and the key, i.e.,
\begin{equation}
 \nabla_{\theta}\mathcal{L}(f_{\theta}(x_t), y) \not \approx \nabla_{\theta}\mathcal{L}(f_{\theta}(x), y),
\end{equation}
and
\begin{equation}
 \nabla_{\theta}\mathcal{L}(f_{\theta}(t(x_b, K_1)), y) \not \approx \nabla_{\theta}\mathcal{L}(f_{\theta}(t(x_b, K_2)), y),
\end{equation}
where $K_1$ and $K_2$ are different keys, and the symbol $\not \approx$ denotes approximately not equal to. Consequently, a model trained by the transformed images works well only when images are transformed under the use of the same transformation and key as those used for training the model. Therefore, the equations
\begin{equation}
 f_{\theta}(x_t) \neq f_{\theta}(x)
\end{equation}
and
\begin{equation}
 f_{\theta}(t(x_b), K_1) \neq f_{\theta}(t(x_b), K_2)
\end{equation}
are satisfied.

\RC{Another notable property of the proposed defense is the low computation cost. The block-wise operation utilized in the proposed defense can be efficiently implemented by vectorized operations and is available for large-scale systems without any noticeable overheads during training/inference. Therefore, the proposed defense has potential for real-world applications including real-time ones.}

\subsection{Key Management and Robustness Against Adaptive Attacks\label{sec:adaptive}}
One of the properties of the proposed transformation is the use of a secret key. \Review{The key can be saved with a provider or can be required by the provider as a parameter for inference as shown in Fig.~\ref{fig:overview}.} Key management and robustness against adaptive attacks are discussed here to evaluate the effectiveness of the proposed defense.

As pointed out in~\cite{Carlini19,Trammer-Arxiv-2020}, adaptive attacks, which are adapted to the specific details of the proposed defense, are necessary in evaluating adversarial defenses to avoid a false sense of security. Optimization-based attack methods require correct gradients of the loss function with respect to the input. Therefore, many defenses make the gradients incorrect by introducing non-differentiable transformation or other obfuscation means such as randomization. These defenses that rely on obfuscated gradients are defeated by adaptive attacks~\cite{AthalyeC018}. One of the reasons adaptive attacks are successful is that useful gradients can be approximated because defensively transformed input is similar to the original input (i.e., $g(x) \approx x$, where $g(\cdot)$ is a defensive transform). In contrast, in the proposed defense, the input is transformed in a systematic way with a secret key, and the resulting input is not similar to the initial input (i.e., $t(x, K) \not\approx x$).

\Review{We carry out the following adaptive attacks to evaluate the proposed defense.} In experiments, the proposed defense will be demonstrated to still maintain robustness against adaptive attacks.

\subsubsection{Inverse Transformation Attack}
\begin{figure}[!t]
\centering
\includegraphics{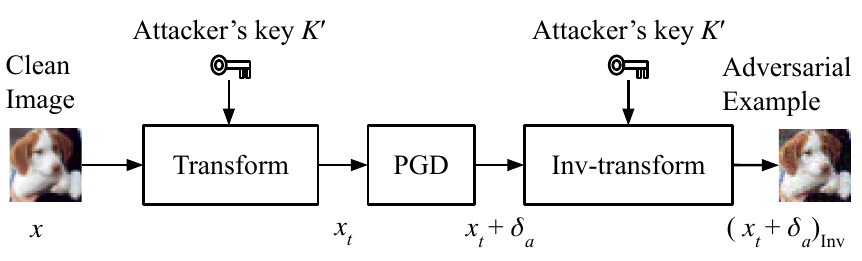}%
\caption{Scenario of adaptive attack with estimated key.\label{fig:adaptive}}
\end{figure}

An adversary may generate adversarial examples by adding noise to transformed images and inverse transform them with an assumed key. To simulate such an attack scenario, we designed an adaptive attack as shown in Fig.~\ref{fig:adaptive}. Since key $K$ is not available to the adversary, it has to be guessed randomly or heuristically for the adversary to carry out the adaptive attack. When an estimated key is close enough to the correct key, the adversary may be able to fool the model. However, we show that searching for a key close to key $K$ is not easy.

\subsubsection{Estimation over Transformation Attack}
\RC{The Estimation over Transformation Attack (EOT) is effective for estimating gradients in adversarial defenses with randomization as explained in~\cite{AthalyeC018}. Instead of taking one step in the direction of gradients $\nabla_{x}f(x)$, we move in the direction of $\sum_{i=1}^{30} \nabla_{x}f(x)$. In other words, we use $30$ keys to generate adversarial examples under a PGD attack.}

\subsubsection{Transferability Attack}
\RC{Since the proposed defensive transformation method is transparent, an attacker can train a substitute model with their key. Then, the attacker generates adversarial examples over the substitute model. We simulate this attack scenario in experiments.}

\section{Experiments\label{sec:experiments}}
To verify the effectiveness of the proposed defense, we ran a number of experiments on different datasets. All the experiments were carried out in PyTorch platform.

\subsection{Datasets}
We used the CIFAR-10~\cite{Krizhevsky09} and ImageNet~\cite{ILSVRC15} datasets.
CIFAR-10 consists of 60,000 color images (dimension of $32 \times 32 \times 3$) with 10 classes (6000 images for each class) where 50,000 images are for training and 10,000 for testing. We utilized a batch size of 128 and live augmentation (random cropping with a padding of 4 and random horizontal flip) on a training set.

\Review{ ImageNet comprises 1.28 million color images for training and 50,000 color images for validation. We progressively resized images during training starting with larger batches of smaller images to smaller batches of larger images. We adapted three phases of training from the DAWNBench top submissions as mentioned in~\cite{Wong-ICLR-2020}. Phases 1 and 2 resized images to 160 and 352 pixels, respectively, and phase 3 used the entire image size from the training set. The augmentation methods used in the experiment were random resizing and cropping (sizes of $128$, $224$, and $288$ respectively for each phase) and random horizontal flip. }

\subsection{Networks}
We utilized deep residual networks~\cite{He16} with 18 layers (ResNet18) for the CIFAR-10 dataset and trained for $200$ epochs with efficient training techniques from the DAWNBench top submissions: cyclic learning rates~\cite{Smith-Arxiv-2017} and mixed-precision training~\cite{Micikevicius-Arxiv-2017}. The parameters of the stochastic gradient descent (SGD) optimizer were a momentum of $0.9$, weight decay of $0.0005$, and maximum learning rate of $0.2$. \Review{ For ImageNet, we used ResNet50 with pre-trained weights. We adapted the training settings from~\cite{Wong-ICLR-2020} with the removal of weight decay regularization from batch normalization layers. The network was trained for 15 epochs in total for the ImageNet dataset. }

\subsection{Attack Settings}
\Review{ We utilized an attack library~\cite{ding2019advertorch} for SPSA, PGD, CW, and EAD attacks, a publicly available implementation of the OnePixel attack, and code from the original authors for NATTACK.\@ }

\Review{ Three black-box attacks, OnePixel, NATTACK, and SPSA, were deployed to evaluate the proposed defense. The OnePixel attack was configured for 10 pixels, 100 iterations, and a population size of 400. For NATTACK, the population size was 300, the sigma was 0.1, the learning rate was 0.02, and 500 iterations were used for the CIFAR-10 dataset, and population sizes of 200 and 200 iterations were used for the ImageNet dataset. SPSA was set up with a delta value of 0.01, a learning rate of 0.01, a batch size of 256, and 100 maximum iterations for CIFAR-10 and a batch size of 128 for ImageNet. }

\Review{ Three white-box attacks, PGD, CW, and EAD, were used to attack the proposed defense. The PGD attack was configured with a step size of 2/255, 50 iterations, and random initialization. Since we focused on untargeted attacks, CW and EAD were configured with a confidence value of $0$, learning rate of $0.01$, binary search steps of $9$, and an initial constant of $0.001$ for $1000$ iterations for CIFAR-10 and $100$ iterations for ImageNet. EAD was set up with the elastic-net (EN) decision rule. }

\subsection{Evaluation Metrics\label{sec:metric}}
\Review{ We used two metrics: accuracy (ACC) and attack success rate (ASR). ACC is given by
\begin{equation}
 \text{ACC} = \left\{ \begin{array}{ll}
 \frac{1}{N}\sum_{i=1}^{N} \mathbbm{1} (f_{\theta}(x_i) = y_i) & (\text{clean})\\
 \frac{1}{N}\sum_{i=1}^{N} \mathbbm{1} (f_{\theta}(x_i + \delta{i}) = y_i) & (\text{attacked}),
 \end{array}
 \right.
\end{equation}
}
and \Review{ ASR is defined as
\begin{equation}
 \text{ASR} = \frac{1}{N}\sum_{i=1}^{N} \mathbbm{1} (f_{\theta}(x_i) = y_i \land f_{\theta}(x_{i} + \delta_{i}) \neq y_i),
\end{equation}
where $N$ is the number of test images, $\mathbbm{1}(\text{condition})$ is one if condition is true, otherwise zero, $\{x_i, y_i\}$ is a test image ($x_i$) with its corresponding label ($y_i$), and $\delta_i$ is its respective adversarial noise depending on a specific attack. }

\subsection{Robustness Against Black-box and White-box Attacks}
\RA{A noise distance $\epsilon$ value of $8/255$ was used in $\ell_\infty$-bounded attacks. Table~\ref{tab:attacks} captures the performance of both the baseline model (standard) and the proposed defense models for different block sizes ($M \in \{2, 4, 8, 16\}$) in terms of clean ACC and ASR.\@ ACC was calculated for the whole test set (10,000 images for CIFAR-10 and 50,000 for ImageNet), and we computed ASR for 1,000 randomly selected images that were correctly classified by the proposed defense models. The models are denoted by their defense method and block size. For example, a model trained by using Pixel Shuffling with a block size of $M = 2$ is indicated as ``Pixel Shuffling ($M = 2$).'' From Table~\ref{tab:attacks}, the results are summarized as follows.}

\subsubsection{CIFAR-10}
\RA{
\begin{itemize}
 \item \textbf{Standard:} Although the baseline (non-protected) model achieved the highest accuracy, it was most vulnerable to all attacks.
 \item \textbf{Pixel Shuffling:} The model with $M = 2$ provided a clean ACC of \SI{94.45}{\percent}, and the worst case ASR was \SI{11.30}{\percent} with SPSA.\@ The model with $M=16$ reduced the clean ACC to \SI{76.22} although the ASRs for all attacks were low. Overall, the model with $M=4$ performed reasonably well whether or not it was under attack.
 \item \textbf{Bit Flipping:} The ACC of the model with $M = 2$ was very close to that of the standard model (i.e., \SI{95.32}{\percent}). However, the ASR was more than \SI{10}{\percent} for OnePixel and SPSA attacks. The models with $M=8$ and $16$ were broken as the ASR was high. For Bit Flipping, the model with $M=4$ achieved the overall best accuracy.
 \item \textbf{FFX Encryption:} Although the ACCs of the models with FFX Encryption were slightly lower, they had better resistance against all of the attacks. Again, for the FFX Encryption defensive transformation, the model with $M = 4$ performed better overall.
\end{itemize}
}

\begin{figure*}[!t]
\centering
\subfloat[Pixel Shuffling]{\includegraphics[width=0.33\linewidth]{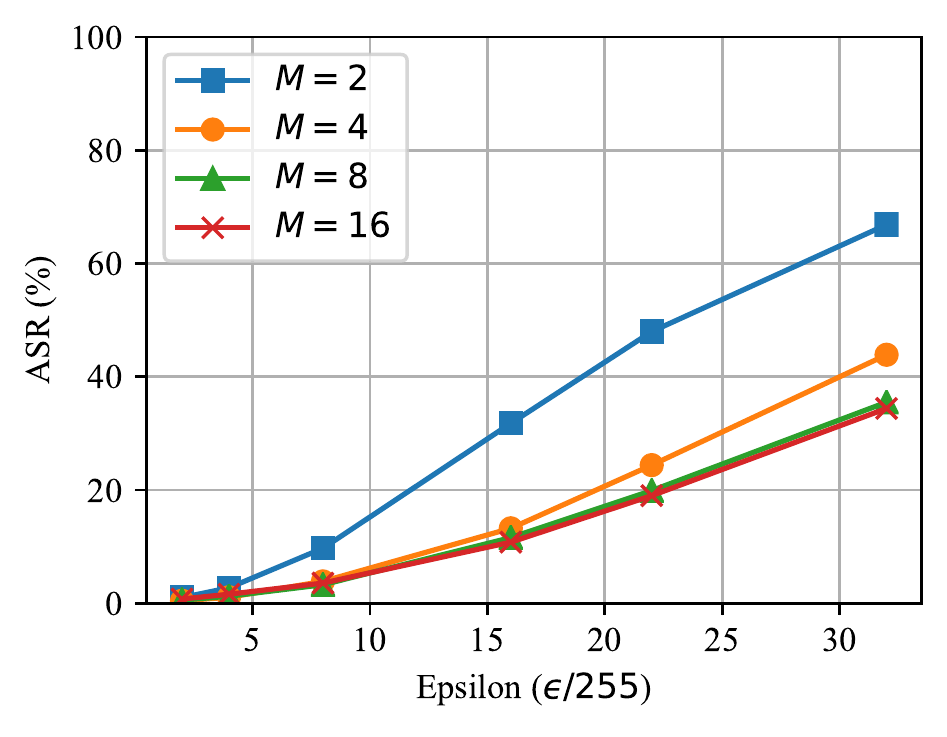}%
\label{fig:shuffle}}
\hfil
\subfloat[Bit Flipping]{\includegraphics[width=0.33\linewidth]{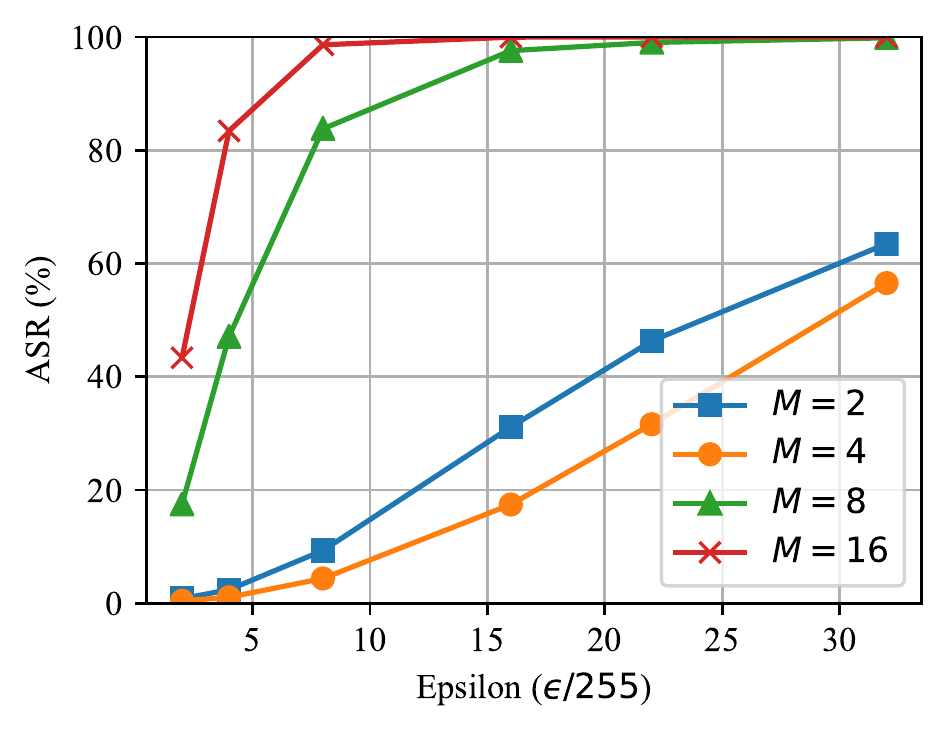}%
\label{fig:flip}}
\hfil
\subfloat[FFX Encryption]{\includegraphics[width=0.33\linewidth]{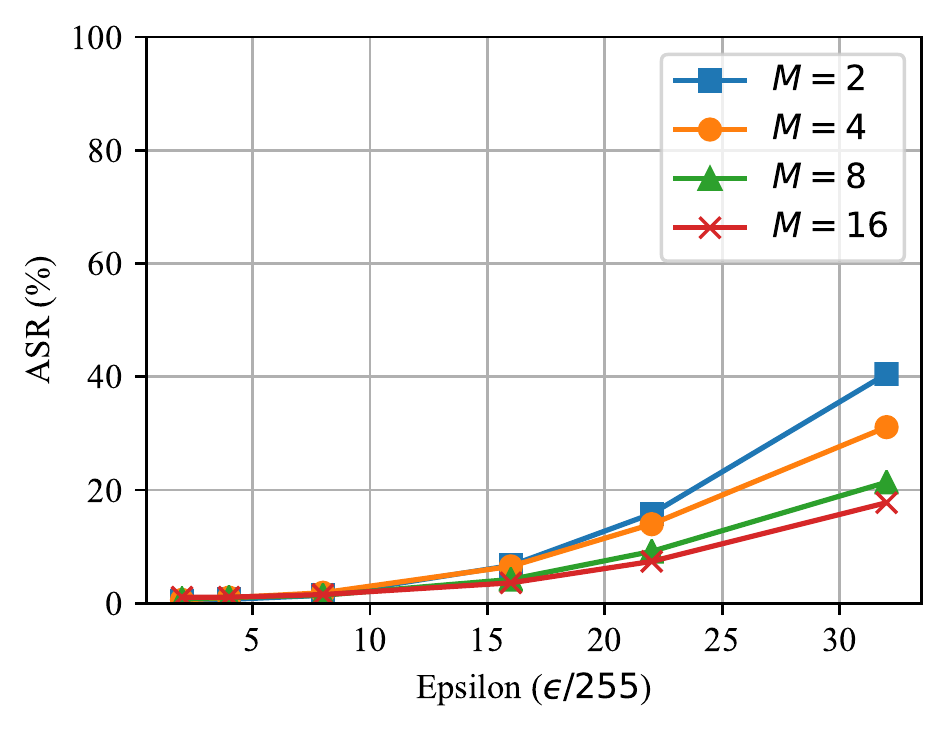}%
\label{fig:encrypt}}
\caption{ASR of proposed defense against PGD attack for CIFAR-10 dataset. ASR was calculated over 10,000 images (whole test set).\label{fig:asr-cifar10}}
\end{figure*}

\RA{In summary, a bigger block size reduced the classification accuracy for Pixel Shuffling and increased the ASR for Bit Flipping. However, for FFX Encryption, although the clean ACC was slightly lower, it provided better defense throughout the attacks (i.e., a lower ASR) for all different block sizes. Our experiments suggest that $M=4$ is the optimal parameter for all three proposed defensive transformations. In addition, we plotted the ASR against noise distance $\epsilon$ (maximum of 32/255) for the whole test set under PGD attack in Fig.~\ref{fig:asr-cifar10}. The ASR for all three transformations increased with respect to bigger $\epsilon$ values. FFX Encryption had a lower ASR throughout all noise levels, and Bit Flipping had a very high ASR for $M = 8$ and $16$.}

\subsubsection{ImageNet}
\begin{figure}[!t]
\centering
\includegraphics[width=\columnwidth]{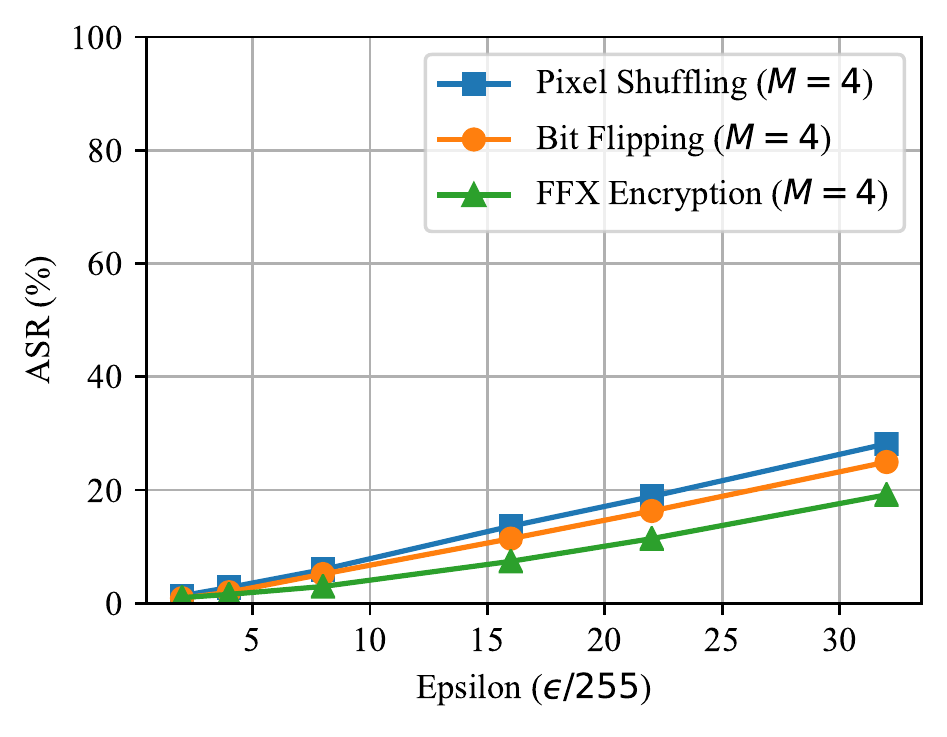}
\caption{ASR of proposed defense against PGD attack for ImageNet dataset. ASR was calculated over 10,000 images randomly selected from validation set.\label{fig:asr-imagenet}}
\end{figure}

\RA{
Since $M=4$ provided overall better results, we used $M=4$ for the ImageNet dataset.
\begin{itemize}
 \item \textbf{Standard:} Similarly, the standard model achieved the highest clean accuracy and ASR for all attacks.
 \item \textbf{Pixel Shuffling:} The ASRs of SPSA and PGD were \SI{6.26}{\percent} and \SI{5.69}{\percent}, respectively, and those of the other attacks were very low.
 \item \textbf{Bit Flipping:} Similarly, the ASRs of SPSA and PGD for Bit Flipping were \SI{6.16}{\percent} and \SI{6.56}{\percent}, respectively, and those of the other attacks were very low.
 \item \textbf{FFX Encryption:} The results show that FFX Encryption provided a lower ASR compared with Pixel Shuffling and Bit Flipping for SPSA and PGD attacks (i.e., \SI{5.77}{\percent} and \SI{3.77}{\percent} respectively). The ASRs of the other attacks were also very low.
\end{itemize}
Notably, the proposed defense achieved almost the same clean accuracy as the standard one (i.e., $\approx$\SI{72}{\percent}), and the ASR was lower than \SI{7}{\percent} for all cases. Figure~\ref{fig:asr-imagenet} shows the performance of the proposed defense with the PGD attack under different noise distances. For the worst-case scenario (i.e., $\epsilon = 32/255$), the ASRs for Pixel Shuffling and Bit Flipping were approximately \SI{28}{\percent} and \SI{25}{\percent}, respectively. In contrast, the ASR of FFX Encryption was less than \SI{20}{\percent}.
}

\robustify\bfseries
\sisetup{table-parse-only,detect-weight=true,detect-inline-weight=text,round-mode=places,round-precision=2}
\begin{table*}[tbp]
 \caption{Clean accuracy (ACC) (\SI{}{\percent}) and Attack Success Rate (ASR) (\SI{}{\percent}) of standard and proposed defense models under different attacks where $\epsilon = 8/255$ for $\ell_\infty$ metric\label{tab:attacks}}
\renewcommand{\arraystretch}{1.5}
 \centering
 \RA{
 \begin{tabular}{l|SS|SSS|SSS}
 \toprule
 \multicolumn{9}{c}{CIFAR-10}\\
 \midrule
 & \multicolumn{2}{c|}{Clean ACC} & \multicolumn{3}{c|}{ASR (Black-box)} & \multicolumn{3}{c}{ASR (White-box)}\\
 \cmidrule{2-9}
   {Model} & {Standard} & {Protected} & {OnePixel ($\ell_0$)} & {NATTACK ($\ell_\infty$)} & {SPSA ($\ell_\infty$)} & {PGD ($\ell_\infty$)} & {CW ($\ell_2$)} & {EAD ($\ell_1$)}\\
 \midrule
   Standard & 95.45 & {--} & 79.90 & 99.9 & 100.0 & 100.0 & 100.0 & 100.0\\
 \midrule
   Pixel Shuffling ($M = 2$) & {\multirow{4}{*}{--}} & \bfseries \num{94.45} & 9.50 & 0.8 & 11.3 & 9.8 & \bfseries \num{0.0} & 0.09\\
   Pixel Shuffling ($M = 4$) & & 91.84 & 5.0 & 0.2 & \bfseries \num{3.0} & \bfseries \num{3.3} & \bfseries \num{0.0} & \bfseries \num{0.0}\\
   Pixel Shuffling ($M = 8$) & & 85.12 & 3.90 & \bfseries \num{0.0} & 3.6 & 4.0 & 0.09 & 0.19\\
   Pixel Shuffling ($M = 16$) & & 76.22 & \bfseries \num{2.80} & 0.2 & 3.37 & 3.42 & \bfseries \num{0.0} & \bfseries \num{0.0}\\
 \midrule
 Bit Flipping ($M = 2$) & {\multirow{4}{*}{--}} & \bfseries \num{95.32} & 10.5 & 0.7 & 10.25 & 9.62 & \bfseries \num{0.0} & 0.18\\
 Bit Flipping ($M = 4$) & & 93.41 & \bfseries \num{5.30} & \bfseries \num{0.2} & \bfseries \num{4.29} & \bfseries \num{4.64} & \bfseries \num{0.0} & \bfseries \num{0.09}\\
 Bit Flipping ($M = 8$) & & 91.54 & 21.40 & 32.5 & 88.14 & 84.34 & 2.26 & 3.11\\
 Bit Flipping ($M = 16$) & & 92.68 & 27.10 & 71.80 & 98.24 & 98.98 & 5.40 & 8.75\\
 \midrule
 FFX Encryption ($M = 2$) & {\multirow{4}{*}{--}} & \bfseries \num{93.67} & 6.3 & \bfseries \num{1.9} & 2.46 & 1.77 & 0.47 & \bfseries \num{0.0}\\
 FFX Encryption ($M = 4$) & & 92.30 & 3.9 & 2.1 & \bfseries \num{2.45} & 1.96 & 0.28 & \bfseries \num{0.0}\\
 FFX Encryption ($M = 8$) & & 91.99 & \bfseries \num{2.0} & 4.2 & 3.57 & \bfseries \num{1.41} & \bfseries \num{0.0} & \bfseries \num{0.0}\\
 FFX Encryption ($M = 16$) & & 91.38 & 3.2 & 6.3 & 6.62 & 1.6 & 0.28 & \bfseries \num{0.0}\\
 \midrule
 \multicolumn{9}{c}{ImageNet}\\
 \midrule
 & \multicolumn{2}{c|}{Clean ACC} & \multicolumn{3}{c|}{ASR (Black-box)} & \multicolumn{3}{c}{ASR (White-box)}\\
 \cmidrule{2-9}
 {Model} & {Standard} & {Protected} & {OnePixel ($\ell_0$)} & {NATTACK ($\ell_\infty$)} & {SPSA ($\ell_\infty$)} & {PGD ($\ell_\infty$)} & {CW ($\ell_2$)} & {EAD ($\ell_1$)}\\
 \midrule
 Standard & \bfseries \num{73.7} & {--} & 13.30 & 97.8 & 99.6 & 100.0 & 100.0 & 100.0\\
 \midrule
 Pixel Shuffling ($M = 4$) & {--} & 72.41 & 1.9 & 0.4 & 6.26 & 5.69 & 0.1 & 0.1\\
 \midrule
 Bit Flipping ($M = 4$) & {--} & 72.63 & 1.2 & \bfseries \num{0.0} & 6.16 & 6.56 & \bfseries \num{0.0} & \bfseries \num{0.0}\\
 \midrule
 FFX Encryption ($M = 4$) & {--} & 72.18 & \bfseries \num{0.7} & 0.6 & \bfseries \num{5.77} & \bfseries \num{3.77} & 0.9 & \bfseries \num{0.0}\\
 \bottomrule
 \end{tabular}
 }
\end{table*}

\subsection{Comparison with State-of-the-art Defenses}
\begin{figure*}[!t]
\centering
\subfloat[Pixel Shuffling]{\includegraphics[width=0.33\linewidth]{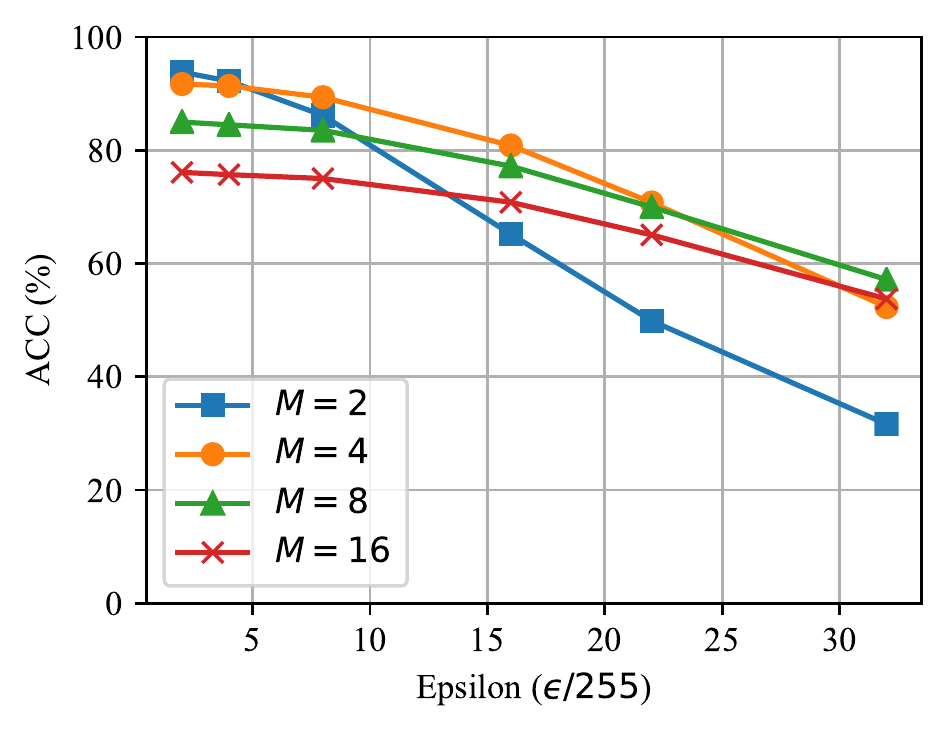}%
\label{fig:acc-shuffle}}
\hfil
\subfloat[Bit Flipping]{\includegraphics[width=0.33\linewidth]{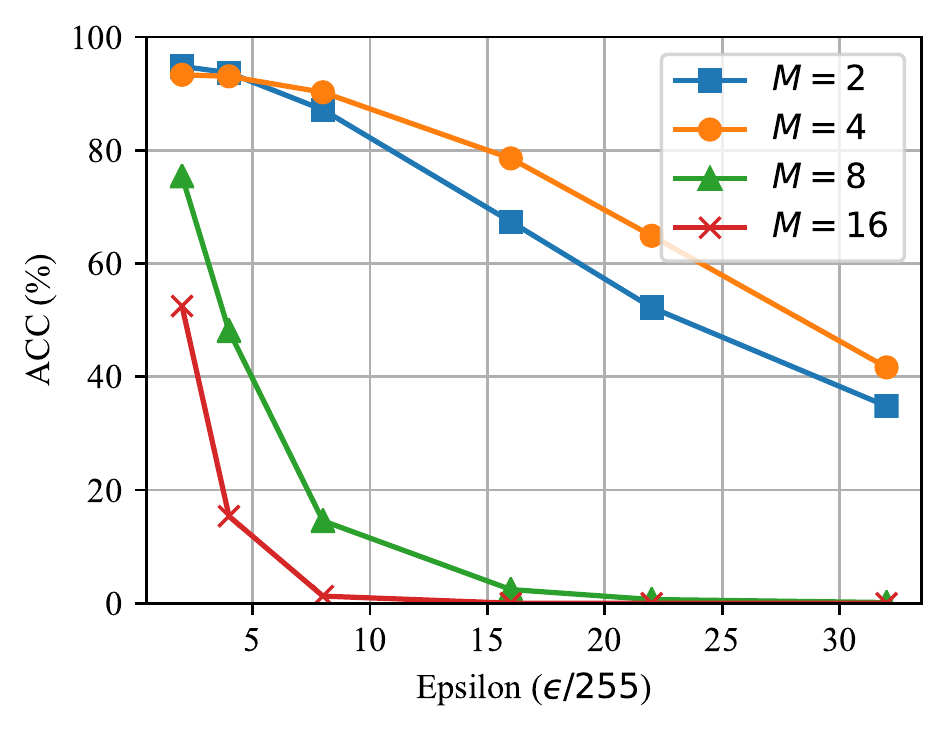}%
\label{fig:acc-flip}}
\hfil
\subfloat[FFX Encryption]{\includegraphics[width=0.33\linewidth]{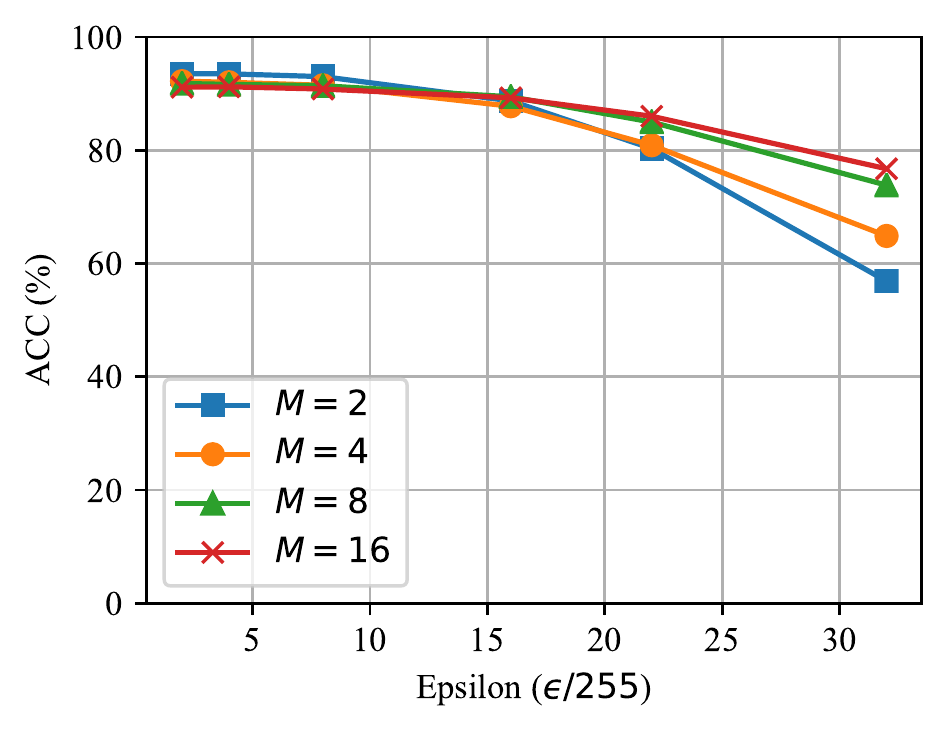}%
\label{fig:acc-encrypt}}
\caption{ACC of proposed defense against PGD attack for CIFAR-10 dataset. ACC was calculated over 10,000 images (whole test set).\label{fig:acc-cifar10}}
\end{figure*}

\RD{First, we compared the accuracy of the proposed defense among the three key-based transformations with different block sizes under PGD attack for the CIFAR-10 dataset. Graphs of accuracy versus perturbation budget $\epsilon$ are shown in Fig.~\ref{fig:acc-cifar10}. When $\epsilon = 8/255$, the model with FFX Encryption ($M = 2$) achieved the highest accuracy (\SI{93.01}{\percent}). As for the worst case $\epsilon$ (i.e., $32/255$), the model with FFX Encryption ($M = 16$) yielded \SI{76.74}{\percent}. Notably, Bit Flipping for $M = 8$ and $16$ reduced the accuracy significantly even for small $\epsilon$ values. However, the models with $M = 4$ provided the overall best accuracy, especially for an $\epsilon$ value of $8/255$ for all three transformations. Therefore, we used the models with $M = 4$ as representatives for comparison with state-of-the-art defenses.}

Most preprocessing-based defenses such as~\cite{Buckman18,Guo17, Xie17, Song17,Samangouei18} were defeated by adaptive attacks due to obfuscated gradients~\cite{AthalyeC018}. Even the most recent state-of-the-art defenses were invalidated by rigorous adaptive attacks~\cite{Trammer-Arxiv-2020}. \RA{To the best of our knowledge, only adversarial training (AT) is repeatedly found effective to defend against adversarial examples. However, AT has been known to be extremely difficult at the ImageNet scale due to the high computation cost~\cite{Kurakin2017}. Recently, ``Fast'' AT was proposed to overcome such difficulty~\cite{Wong-ICLR-2020}. We compared the proposed defense models with the latest efficient AT (i.e., Fast AT)~\cite{Wong-ICLR-2020} as a baseline defense, a recent feature scattering-based approach (FS)~\cite{2019-NIPS-Zhang}}, and another key-based defense using standard random permutation (SRP)~\cite{Taran-ECCV-2018} in terms of accuracy, whether or not the model was under PGD attack with various perturbation budgets. We exclude defenses that are already broken or that have a very low clean accuracy from comparison.

\subsubsection{CIFAR-10}
\begin{figure}[!t]
\centering
\includegraphics[width=\columnwidth]{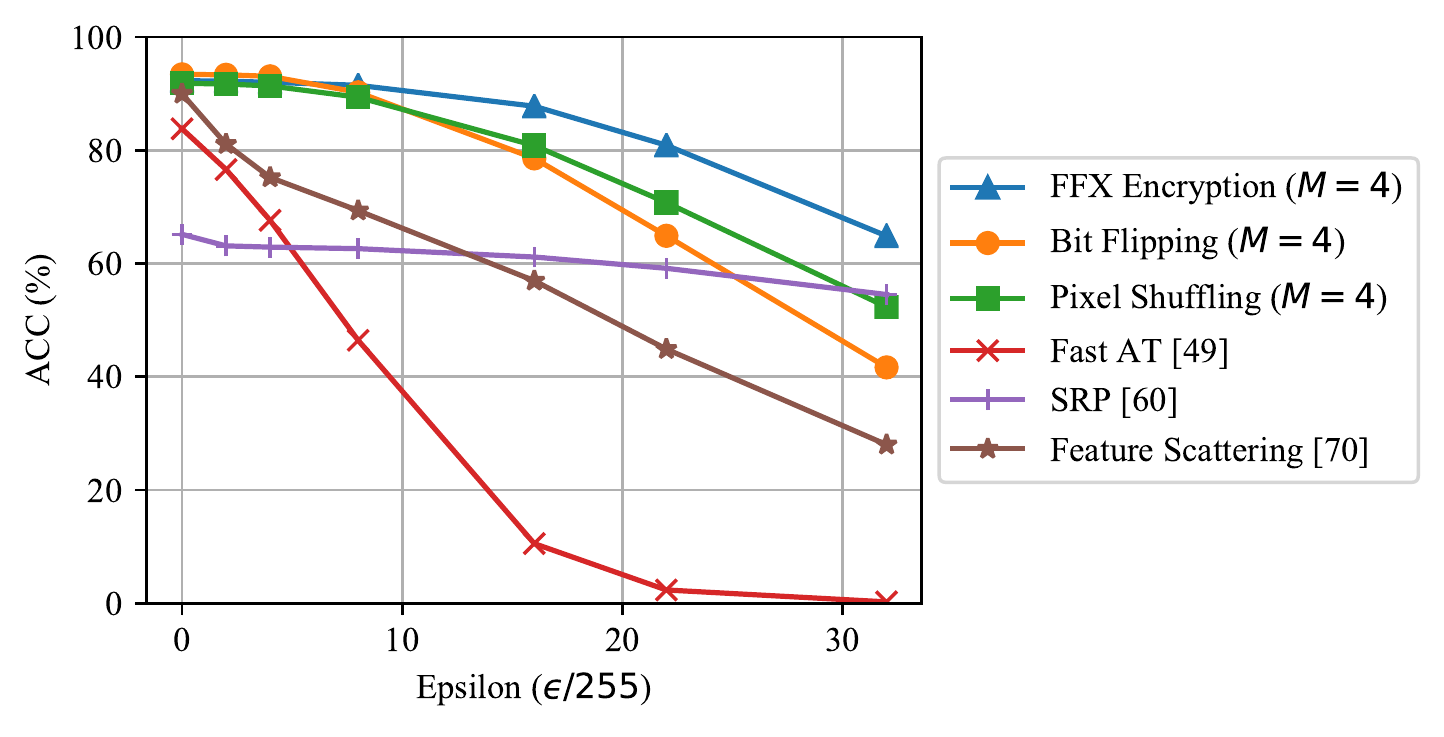}
\caption{Comparison of proposed defense with state-of-the-art defenses in terms of ACC under PGD attack for CIFAR-10 dataset. ACC was calculated over 10,000 images (whole test set).\label{fig:cifar10-compare}}
\end{figure}

\RD{Figure~\ref{fig:cifar10-compare} shows the performance of the proposed defense models with $M = 4$ compared with Fast AT~\cite{Wong-ICLR-2020}, FS~\cite{2019-NIPS-Zhang}, and SRP~\cite{Taran-ECCV-2018}. In terms of clean accuracy, the model with Bit Flipping ($M = 4$) achieved the highest accuracy (i.e., \SI{93.41}{\percent}), while Fast AT was \SI{83.80}{\percent}, FS was \SI{89.98}{\percent}, and SRP was \SI{65.16}{\percent}. When the noise distance was 8/255, the model with FFX Encryption ($M = 4$) outperformed all of the methods, achieving \SI{91.48}{\percent} compared with Fast AT (\SI{46.44}{\percent}), FS (\SI{69.35}{\percent}), and SRP (\SI{62.63}{\percent}). When the perturbation budget was increased to 32/255, the model with FFX Encryption ($M = 4$) still provided the highest accuracy (\SI{64.86}{\percent}). Overall, all of the models with the proposed transformations outperformed state-of-the-art defenses at any given perturbation budget.}

\subsubsection{ImageNet}
\begin{figure}[!t]
\centering
\includegraphics[width=\columnwidth]{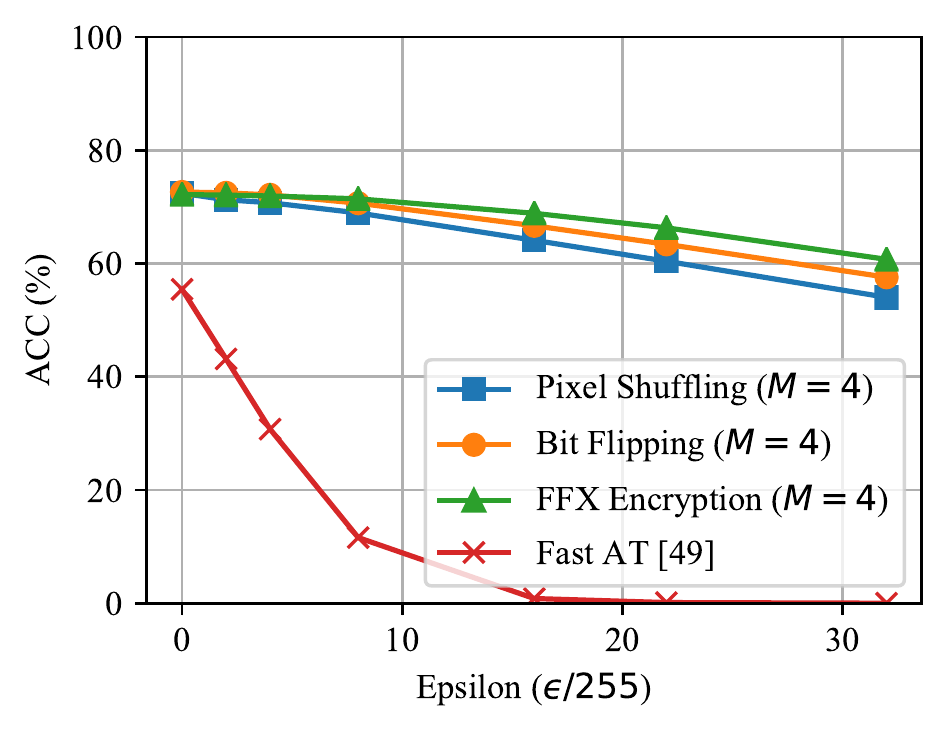}
\caption{Comparison of proposed defense with state-of-the-art defenses in terms of ACC under PGD attack for ImageNet dataset. ACC was calculated over 50,000 images (whole validation set).\label{fig:imagenet-compare}}
\end{figure}

\RD{In a similar fashion, we conducted the PGD attack with different perturbation budgets to confirm the effectiveness of the proposed defense. The accuracy of SRP~\cite{Taran-ECCV-2018} for ImageNet was \SI{9.99}{\percent}; therefore, we excluded SRP from comparison. Moreover, FS~\cite{2019-NIPS-Zhang} is not available for the ImageNet dataset. Therefore, we compared the proposed defense with the Fast AT~\cite{Wong-ICLR-2020} released by the original authors, which was trained with an $\epsilon$ value of 4/255. Figure~\ref{fig:imagenet-compare} shows the performance comparison under the PGD attack with different $\epsilon$ values in terms of ACC.\@ The model with FFX Encryption outperformed all other methods for any given perturbation budget. In the literature, there is no defense that can maintain clean accuracy close to the standard one at the ImageNet scale. We are the first to achieve the closest clean accuracy as well as a high accuracy under the attacks even on the ImageNet dataset.}

\subsection{Robustness Against Adaptive Attacks}
\RA{Without the correct key or a near-correct key, conventional attacks will not work on the proposed defense with a secret key. Therefore, we assume an attacker may estimate the correct key $K$ randomly or heuristically. Once $K$ was estimated, we ran the PGD attack by using the estimated key $K'$ since PGD is one of the strongest adversaries, and the proposed defense was confirmed effective when the key was correct. Apart from key estimation methods, we also deployed the adaptive attacks described in~\ref{sec:adaptive} to evaluate the proposed defense.}

\subsubsection{Random Key Estimation Approach}
One of the ways of estimating key $K$ is to randomly search for a key. \RC{As in black-box settings, the attacker may query the model with their key. We allow the attacker to query the model for a maximum of 20,000 queries. In other words, the attacker uses a single image and a key $K'$ at a time to query the model. When the model makes the correct prediction for the test image with respect to the key $K'$, the attacker stops the random search and uses $K'$ to generate adversarial examples. While considering the worst-case scenario, we also assume the attacker has the weights of the model (white-box) and can use a batch of images to test a key $K'$ over the average accuracy.}

\RC{
The key space can be varied depending on the number of pixels in a block $p_b$. The key space of Pixel Shuffling is given by
\begin{equation}
 \mathcal{K}_{\text{shuffling}}(p_b) = p_b!.
\end{equation}
For Bit Flipping and FFX Encryption, \SI{50}{\percent} of the pixels in each block are inversed/encrypted, and the key controls which pixels are inversed/encrypted. Therefore, their key spaces are the same and written as
\begin{equation}
 \mathcal{K}_{\text{flipping/encryption}}(p_b) = \begin{pmatrix} p_b \\ \frac{p_b}{2} \end{pmatrix} = \frac{p_b!}{(p_b / 2)! \cdot (p_b / 2)!}. \label{eq:ffx-key}
\end{equation}
}


\subsubsection{Heuristic Key Estimation Approach}
\RC{As in white-box settings, we assume the attacker knows the model weights and inner workings of the defense algorithm. In this case, instead of trying a key randomly, key $K$ may be estimated by using a heuristic approach. In other words, $K$ is not directly estimated, but the transformation pattern caused by $K$ is estimated. A key is used to generate a random permutation vector $v = (v_0, v_1,\dots,v_{p_b - 1})$ for Pixel Shuffling and a random binary vector $r = (r_0, r_1, \dots, r_{p_b - 1})$ for Bit Flipping and FFX Encryption. Therefore, the adversary can modify $v$ or $r$ by using the average accuracy over a batch of images as a guide to carry out an adaptive attack as follows (see Algorithm~\ref{algo:key1}).}

\begin{enumerate}
\item Initialize a permutation vector $v$ (for Pixel Shuffling) or a binary vector $r$ (for Bit Flipping/FFX Encryption) with a random key $K'$.
\item Calculate the accuracy of the model over a batch of images.
\item Repeatedly swap two values in $v$: $v_i$ and $v_{j}$ (for Pixel Shuffling) or in $r$: $r_i$ and $r_{j}$ (for Bit Flipping/FFX Encryption) for $T$ rounds if the accuracy improves.
\item Return the tuned $v$ (for Pixel Shuffling) or $r$ (for Bit Flipping/FFX Encryption) to proceed with the adaptive attack.
\end{enumerate}

\begin{algorithm}
\caption{Heuristic Key Estimation Approach\label{algo:key1}}
 \begin{algorithmic}
 \renewcommand{\algorithmicrequire}{\textbf{Input:}}
 \renewcommand{\algorithmicensure}{\textbf{Output:}}
 \REQUIRE{A batch of images}
 \ENSURE{$v$ or $r$}
 \STATE{Initialize $v$ or $r$ with a random key $K'$}
 \STATE{accuracy $\leftarrow$ Calculate the accuracy of the model}
 \FOR{$t \leftarrow 1\dots T$}
 \FOR{$i \leftarrow 0, \dots, p_b - 1 $}
 \FOR{$j \leftarrow i+1, \dots, p_b - 1$}
 \IF{accuracy improves}
 \STATE{Swap $v_i$ and $v_j$ for Pixel Shuffling}
 \STATE{or}
 \STATE{Swap $r_i$ and $r_j$ for Bit Flipping/FFX Encryption}
 \ENDIF
 \ENDFOR
 \ENDFOR
 \ENDFOR
 \end{algorithmic}
\end{algorithm}

\RC{We implemented the key search approaches on the CIFAR-10 dataset with a batch size of 128 and parameter $T = 10$. A note on FFX Encryption is that password $P$ does not matter since the length of FFX encryption is fixed (i.e., $3$). Therefore, the attacker can assume any password during the attack. Table~\ref{tab:adaptive} summarizes the results of the key search attacks. In all transformations, random key search approaches (either by a single image or batch of images) did not guarantee that a close-enough key was found since ASR was very low. For the heuristic approach, ASR was \SI{77.76}{\percent} for Bit Flipping ($M = 4$), \SI{3.7}{\percent} for Pixel Shuffling ($M = 4$), and \SI{3.27}{\percent} for FFX Encryption ($M = 4$). Although the ASR for Bit Flipping was high, this type of attack is only possible when the model weights are available to the attacker. However, Pixel Shuffling and FFX Encryption were still resistant to such attacks. Moreover, one key belongs to one model only and, therefore, the attacker cannot generalize the attack.}

\subsubsection{Inverse Transformation Attack}
In Fig.~\ref{fig:example}, examples of adversarial and adaptive adversarial examples are illustrated under the PGD attack for each algorithm, where key $K$ was estimated by using the heuristic approach (see Algorithm~\ref{algo:key1}) with $T=10$. For FFX Encryption, the visibility of the adaptive adversarial example was heavily changed compared with Pixel Shuffling and Bit Flipping. \RD{In other words, the perturbations were clearly perceptive, and valid adversarial examples were not found under this type of attack for FFX Encryption. Therefore, we do not report the result of this adaptive attack for FFX Encryption in Table~\ref{tab:adaptive}. Since the estimated key was not good enough, the ASR was still very low for both Pixel Shuffling and Bit Flipping.}

\newcolumntype{P}[1]{>{\centering\arraybackslash}p{#1}}
\newcolumntype{M}[1]{>{\centering\arraybackslash}m{#1}}

\begin{figure*}[!t]
 \centering
 \begin{tabular}{M{1.0in}M{1.5in}M{1.5in}M{1.5in}}
 {Model} & {Original Image } & {Adversarial Example} & {Adaptive Adversarial Example}\\
 {} & {$x$} & {$x + \delta$} & {${(x_t + \delta_a)}_{\text{Inv}}$}\\
 \addlinespace
 Pixel Shuffling ($M = 4$) & \includegraphics[width=0.33\columnwidth, valign=c]{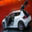} & \includegraphics[width=0.33\columnwidth, valign=c]{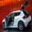} & \includegraphics[width=0.33\columnwidth, valign=c]{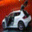}\\
 \addlinespace[0.2em]
 Bit Flipping ($M = 4$) & \includegraphics[width=0.33\columnwidth, valign=c]{car} & \includegraphics[width=0.33\columnwidth, valign=c]{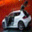} & \includegraphics[width=0.33\columnwidth, valign=c]{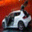}\\
 \addlinespace[0.2em]
 FFX Encryption ($M = 4$) & \includegraphics[width=0.33\columnwidth, valign=c]{car} & \includegraphics[width=0.33\columnwidth, valign=c]{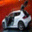} & \includegraphics[width=0.33\columnwidth, valign=c]{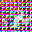}\\
 \end{tabular}
\caption{Example of original test image, adversarial examples, and adaptive adversarial examples generated with estimated key for PGD ($\ell_\infty$) with $\epsilon = 8/255$, where adaptive adversarial example includes severe distortion for FFX Encryption.\label{fig:example}}
\end{figure*}

\subsubsection{Estimation over Transformation Attack}
\RC{The results for the EOT attack are summarized in Table~\ref{tab:adaptive}. From the experiments, the ASR was also very low (less than \SI{2}{\percent}) for Pixel Shuffling and Bit Flipping and $\approx$\SI{5}{\percent} for FFX Encryption. Therefore, the proposed defense was still resistant against such adversarial examples.}

\subsubsection{Transferability Attack}
\RC{We simulated this attack scenario, and the results are presented in Table~\ref{tab:adaptive}. The ASR was \SI{5.07}{\percent} for Pixel Shuffling, \SI{1.49}{\percent} for Bit Flipping, and \SI{6.41}{\percent} for FFX Encryption. The results suggest that the proposed method can still defend against adversarial examples under this type of attack.}

\robustify\bfseries
\sisetup{table-parse-only,detect-weight=true,detect-inline-weight=text,round-mode=places,round-precision=2}
\begin{table*}[tbp]
 \caption{Attack Success Rate (ASR) (\SI{}{\percent}) of adaptive attacks for CIFAR-10 dataset\label{tab:adaptive}}
\renewcommand{\arraystretch}{1.5}
 \centering
 \RC{
 \begin{tabular}{l|SSS|S|S|S}
 \toprule
 & \multicolumn{3}{c|}{Key Search} & & &\\
 & \multicolumn{2}{c}{Random} & {Heuristic} & {Inverse} & & {Transferability}\\
 {Model} & {Single} & {Batch} & & {Transformation} & {EOT} & {Attack}\\
 \midrule
 Pixel Shuffling ($M = 4$) & 3.8 & 4.45 & 3.70 & 3.87 & 1.7 & 5.07\\
 Bit Flipping ($M = 4$) & 3.5 & 3.82 & 77.76 & 78.22 & 1.58 & 1.49\\
 FFX Encryption ($M = 4$) & 1.7 & 1.23 & 3.27 & {--} & 5.24 & 6.41\\
 \bottomrule
 \end{tabular}
 }
\end{table*}

\section{Conclusion\label{sec:conclusion}}
In this paper, we proposed a novel block-wise image transformation as a preprocessing defense method, where both input images and test ones are preprocessed by using the proposed transformation with a key. To realize the proposed transformation, we developed three algorithms: Pixel Shuffling, Bit Flipping, and FFX Encryption. The results showed that the proposed defense was robust against conventional threat models under various metrics ($\ell_\infty, \ell_2, \ell_1, \ell_0$), achieving more than \SI{90}{\percent} accuracy for both clean images and adversarial examples. In addition, we also conducted various adaptive attacks to further evaluate the effectiveness of the proposed defense. Under PGD attack with different perturbation budgets, the proposed defense outperformed the state-of-the art adversarial defenses with the CIFAR-10 and ImageNet datasets. Moreover, the proposed defense was confirmed to bring robust accuracy close to non-robust accuracy for both the CIFAR-10 and ImageNet datasets for the first time.

\ifCLASSOPTIONcaptionsoff
 \newpage
\fi

\bibliographystyle{IEEEtran}
\bibliography{ref}

\begin{thebibliography}{10}
\providecommand{\url}[1]{#1}
\csname url@samestyle\endcsname
\providecommand{\newblock}{\relax}
\providecommand{\bibinfo}[2]{#2}
\providecommand{\BIBentrySTDinterwordspacing}{\spaceskip=0pt\relax}
\providecommand{\BIBentryALTinterwordstretchfactor}{4}
\providecommand{\BIBentryALTinterwordspacing}{\spaceskip=\fontdimen2\font plus
\BIBentryALTinterwordstretchfactor\fontdimen3\font minus
  \fontdimen4\font\relax}
\providecommand{\BIBforeignlanguage}[2]{{%
\expandafter\ifx\csname l@#1\endcsname\relax
\typeout{** WARNING: IEEEtran.bst: No hyphenation pattern has been}%
\typeout{** loaded for the language `#1'. Using the pattern for}%
\typeout{** the default language instead.}%
\else
\language=\csname l@#1\endcsname
\fi
#2}}
\providecommand{\BIBdecl}{\relax}
\BIBdecl

\bibitem{fredrikson2015model}
M.~Fredrikson, S.~Jha, and T.~Ristenpart, ``Model inversion attacks that
  exploit confidence information and basic countermeasures,'' in
  \emph{Proceedings of the 22nd ACM SIGSAC Conference on Computer and
  Communications Security}.\hskip 1em plus 0.5em minus 0.4em\relax ACM, 2015,
  pp. 1322--1333.

\bibitem{shokri2017membership}
R.~Shokri, M.~Stronati, C.~Song, and V.~Shmatikov, ``Membership inference
  attacks against machine learning models,'' in \emph{2017 IEEE Symposium on
  Security and Privacy (SP)}.\hskip 1em plus 0.5em minus 0.4em\relax IEEE,
  2017, pp. 3--18.

\bibitem{Szegedy14}
\BIBentryALTinterwordspacing
C.~Szegedy, W.~Zaremba, I.~Sutskever, J.~Bruna, D.~Erhan, I.~Goodfellow, and
  R.~Fergus, ``Intriguing properties of neural networks,'' in
  \emph{International Conference on Learning Representations}, 2014. [Online].
  Available: \url{http://arxiv.org/abs/1312.6199}
\BIBentrySTDinterwordspacing

\bibitem{Biggio13}
B.~Biggio, I.~Corona, D.~Maiorca, B.~Nelson, N.~{\v{S}}rndi{\'c}, P.~Laskov,
  G.~Giacinto, and F.~Roli, ``Evasion attacks against machine learning at test
  time,'' in \emph{Joint European conference on machine learning and knowledge
  discovery in databases}.\hskip 1em plus 0.5em minus 0.4em\relax Springer,
  2013, pp. 387--402.

\bibitem{Goodfellow15}
I.~J. Goodfellow, J.~Shlens, and C.~Szegedy, ``Explaining and harnessing
  adversarial examples,'' in \emph{3rd International Conference on Learning
  Representations, {ICLR} 2015, San Diego, CA, USA, May 7-9, 2015, Conference
  Track Proceedings}, 2015.

\bibitem{Kurakin2017}
A.~Kurakin, I.~J. Goodfellow, and S.~Bengio, ``Adversarial machine learning at
  scale,'' in \emph{5th International Conference on Learning Representations,
  {ICLR} 2017, Toulon, France, April 24-26, 2017, Conference Track
  Proceedings}, 2017.

\bibitem{Dezfooli16}
S.~Moosavi{-}Dezfooli, A.~Fawzi, and P.~Frossard, ``Deepfool: {A} simple and
  accurate method to fool deep neural networks,'' in \emph{2016 {IEEE}
  Conference on Computer Vision and Pattern Recognition, {CVPR} 2016, Las
  Vegas, NV, USA, June 27-30, 2016}, 2016, pp. 2574--2582.

\bibitem{Carlini017}
N.~Carlini and D.~A. Wagner, ``Towards evaluating the robustness of neural
  networks,'' in \emph{2017 {IEEE} Symposium on Security and Privacy, {SP}
  2017, San Jose, CA, USA, May 22-26, 2017}, 2017, pp. 39--57.

\bibitem{Madry18}
\BIBentryALTinterwordspacing
A.~Madry, A.~Makelov, L.~Schmidt, D.~Tsipras, and A.~Vladu, ``Towards deep
  learning models resistant to adversarial attacks,'' in \emph{6th
  International Conference on Learning Representations, {ICLR} 2018, Vancouver,
  BC, Canada, April 30 - May 3, 2018, Conference Track Proceedings}, 2018.
  [Online]. Available: \url{https://openreview.net/forum?id=rJzIBfZAb}
\BIBentrySTDinterwordspacing

\bibitem{engstrom2017rotation}
\BIBentryALTinterwordspacing
L.~Engstrom, D.~Tsipras, L.~Schmidt, and A.~Madry, ``A rotation and a
  translation suffice: Fooling cnns with simple transformations,''
  \emph{arXiv:1712.02779}, 2017. [Online]. Available:
  \url{http://arxiv.org/abs/1712.02779}
\BIBentrySTDinterwordspacing

\bibitem{gilmer2018motivating}
\BIBentryALTinterwordspacing
J.~Gilmer, R.~P. Adams, I.~J. Goodfellow, D.~Andersen, and G.~E. Dahl,
  ``Motivating the rules of the game for adversarial example research,''
  \emph{arXiv:1807.06732}, 2018. [Online]. Available:
  \url{http://arxiv.org/abs/1807.06732}
\BIBentrySTDinterwordspacing

\bibitem{Eykholt18}
K.~Eykholt, I.~Evtimov, E.~Fernandes, B.~Li, A.~Rahmati, C.~Xiao, A.~Prakash,
  T.~Kohno, and D.~Song, ``Robust physical-world attacks on deep learning
  visual classification,'' in \emph{2018 {IEEE} Conference on Computer Vision
  and Pattern Recognition, {CVPR} 2018, Salt Lake City, UT, USA, June 18-22,
  2018}, 2018, pp. 1625--1634.

\bibitem{Athalye18}
\BIBentryALTinterwordspacing
A.~Athalye, L.~Engstrom, A.~Ilyas, and K.~Kwok, ``Synthesizing robust
  adversarial examples,'' in \emph{Proceedings of the 35th International
  Conference on Machine Learning, {ICML} 2018, Stockholmsm{\"{a}}ssan,
  Stockholm, Sweden, July 10-15, 2018}, 2018, pp. 284--293. [Online].
  Available: \url{http://proceedings.mlr.press/v80/athalye18b.html}
\BIBentrySTDinterwordspacing

\bibitem{KurakinGB17a}
A.~Kurakin, I.~J. Goodfellow, and S.~Bengio, ``Adversarial examples in the
  physical world,'' in \emph{5th International Conference on Learning
  Representations, {ICLR} 2017, Toulon, France, April 24-26, 2017, Workshop
  Track Proceedings}, 2017.

\bibitem{Papernot17}
\BIBentryALTinterwordspacing
N.~Papernot, P.~D. McDaniel, I.~J. Goodfellow, S.~Jha, Z.~B. Celik, and
  A.~Swami, ``Practical black-box attacks against machine learning,'' in
  \emph{Proceedings of the 2017 {ACM} on Asia Conference on Computer and
  Communications Security, AsiaCCS 2017, Abu Dhabi, United Arab Emirates, April
  2-6, 2017}, 2017, pp. 506--519. [Online]. Available:
  \url{https://doi.org/10.1145/3052973.3053009}
\BIBentrySTDinterwordspacing

\bibitem{Sharif2016}
\BIBentryALTinterwordspacing
M.~Sharif, S.~Bhagavatula, L.~Bauer, and M.~K. Reiter, ``Accessorize to a
  crime: Real and stealthy attacks on state-of-the-art face recognition,'' in
  \emph{Proceedings of the 2016 {ACM} {SIGSAC} Conference on Computer and
  Communications Security, Vienna, Austria, October 24-28, 2016}, 2016, pp.
  1528--1540. [Online]. Available:
  \url{https://doi.org/10.1145/2976749.2978392}
\BIBentrySTDinterwordspacing

\bibitem{Carlini19}
\BIBentryALTinterwordspacing
N.~Carlini, A.~Athalye, N.~Papernot, W.~Brendel, J.~Rauber, D.~Tsipras, I.~J.
  Goodfellow, A.~Madry, and A.~Kurakin, ``On evaluating adversarial
  robustness,'' \emph{arXiv:1902.06705}, 2019. [Online]. Available:
  \url{http://arxiv.org/abs/1902.06705}
\BIBentrySTDinterwordspacing

\bibitem{AthalyeC018}
A.~Athalye, N.~Carlini, and D.~A. Wagner, ``Obfuscated gradients give a false
  sense of security: Circumventing defenses to adversarial examples,'' in
  \emph{ICML}, 2018, pp. 274--283.

\bibitem{Trammer-Arxiv-2020}
\BIBentryALTinterwordspacing
F.~Tram{\`{e}}r, N.~Carlini, W.~Brendel, and A.~Madry, ``On adaptive attacks to
  adversarial example defenses,'' \emph{arXiv:2002.08347}, 2020. [Online].
  Available: \url{https://arxiv.org/abs/2002.08347}
\BIBentrySTDinterwordspacing

\bibitem{Chuman-TIFS-2019}
T.~Chuman, W.~Sirichotedumrong, and H.~Kiya, ``Encryption-then-compression
  systems using grayscale-based image encryption for jpeg images,'' \emph{IEEE
  Transactions on Information Forensics and Security}, vol.~14, no.~6, pp.
  1515--1525, June 2019.

\bibitem{Warit-APSIPAT-2019}
W.~Sirichotedumrong and H.~Kiya, ``Grayscale-based block scrambling image
  encryption using ycbcr color space for encryption-then-compression systems,''
  \emph{APSIPA Transactions on Signal and Information Processing}, vol.~8,
  2019.

\bibitem{2019-ICIP-Warit}
W.~Sirichotedumrong, T.~Maekawa, Y.~Kinoshita, and H.~Kiya,
  ``Privacy-preserving deep neural networks with pixel-based image encryption
  considering data augmentation in the encrypted domain,'' in \emph{2019 IEEE
  International Conference on Image Processing (ICIP)}.\hskip 1em plus 0.5em
  minus 0.4em\relax IEEE, 2019, pp. 674--678.

\bibitem{Warit-Access-2019}
W.~Sirichotedumrong, Y.~Kinoshita, and H.~Kiya, ``Pixel-based image encryption
  without key management for privacy-preserving deep neural networks,''
  \emph{IEEE Access}, vol.~7, pp. 177\,844--177\,855, 2019.

\bibitem{madono2020block}
\BIBentryALTinterwordspacing
K.~Madono, M.~Tanaka, M.~Onishi, and T.~Ogawa, ``Block-wise scrambled image
  recognition using adaptation network,'' \emph{arXiv:2001.07761}, 2020.
  [Online]. Available: \url{https://arxiv.org/abs/2001.07761}
\BIBentrySTDinterwordspacing

\bibitem{Tanaka-ICCETW-2018}
M.~Tanaka, ``Learnable image encryption,'' in \emph{2018 IEEE International
  Conference on Consumer Electronics-Taiwan (ICCE-TW)}.\hskip 1em plus 0.5em
  minus 0.4em\relax IEEE, 2018, pp. 1--2.

\bibitem{Kurihara-IEICE-2017}
K.~Kurihara, S.~Imaizumi, S.~Shiota, and H.~Kiya, ``An
  encryption-then-compression system for lossless image compression
  standards,'' \emph{IEICE transactions on information and systems}, vol. 100,
  no.~1, pp. 52--56, 2017.

\bibitem{2019-JMLR-Azulay}
\BIBentryALTinterwordspacing
A.~Azulay and Y.~Weiss, ``Why do deep convolutional networks generalize so
  poorly to small image transformations?'' \emph{Journal of Machine Learning
  Research}, vol.~20, no. 184, pp. 1--25, 2019. [Online]. Available:
  \url{http://jmlr.org/papers/v20/19-519.html}
\BIBentrySTDinterwordspacing

\bibitem{Maung-ICIP-2020}
\BIBentryALTinterwordspacing
M.~AprilPyone and H.~Kiya, ``Encryption inspired adversarial defense for visual
  classification,'' \emph{arXiv:2005.07998}, 2020. [Online]. Available:
  \url{https://arxiv.org/abs/2005.07998}
\BIBentrySTDinterwordspacing

\bibitem{barreno2010security}
M.~Barreno, B.~Nelson, A.~D. Joseph, and J.~D. Tygar, ``The security of machine
  learning,'' \emph{Machine Learning}, vol.~81, no.~2, pp. 121--148, 2010.

\bibitem{shafahi2018poison}
A.~Shafahi, W.~R. Huang, M.~Najibi, O.~Suciu, C.~Studer, T.~Dumitras, and
  T.~Goldstein, ``Poison frogs! targeted clean-label poisoning attacks on
  neural networks,'' in \emph{Advances in Neural Information Processing
  Systems}, 2018, pp. 6103--6113.

\bibitem{Chen-AAAI-2018}
P.~Chen, Y.~Sharma, H.~Zhang, J.~Yi, and C.~Hsieh, ``{EAD:} elastic-net attacks
  to deep neural networks via adversarial examples,'' in \emph{Proceedings of
  the Thirty-Second {AAAI} Conference on Artificial Intelligence, (AAAI-18),
  the 30th innovative Applications of Artificial Intelligence (IAAI-18), and
  the 8th {AAAI} Symposium on Educational Advances in Artificial Intelligence
  (EAAI-18), New Orleans, Louisiana, USA, February 2-7, 2018}, S.~A. McIlraith
  and K.~Q. Weinberger, Eds.\hskip 1em plus 0.5em minus 0.4em\relax {AAAI}
  Press, 2018, pp. 10--17.

\bibitem{dong2018boosting}
Y.~Dong, F.~Liao, T.~Pang, H.~Su, J.~Zhu, X.~Hu, and J.~Li, ``Boosting
  adversarial attacks with momentum,'' in \emph{Proceedings of the IEEE
  conference on computer vision and pattern recognition}, 2018, pp. 9185--9193.

\bibitem{dong2019evading}
Y.~Dong, T.~Pang, H.~Su, and J.~Zhu, ``Evading defenses to transferable
  adversarial examples by translation-invariant attacks,'' in \emph{Proceedings
  of the IEEE Conference on Computer Vision and Pattern Recognition}, 2019, pp.
  4312--4321.

\bibitem{xie2019improving}
C.~Xie, Z.~Zhang, Y.~Zhou, S.~Bai, J.~Wang, Z.~Ren, and A.~L. Yuille,
  ``Improving transferability of adversarial examples with input diversity,''
  in \emph{Proceedings of the IEEE Conference on Computer Vision and Pattern
  Recognition}, 2019, pp. 2730--2739.

\bibitem{chen2017zoo}
P.-Y. Chen, H.~Zhang, Y.~Sharma, J.~Yi, and C.-J. Hsieh, ``Zoo: Zeroth order
  optimization based black-box attacks to deep neural networks without training
  substitute models,'' in \emph{Proceedings of the 10th ACM Workshop on
  Artificial Intelligence and Security}, 2017, pp. 15--26.

\bibitem{ilyas2018black}
\BIBentryALTinterwordspacing
A.~Ilyas, L.~Engstrom, A.~Athalye, and J.~Lin, ``Black-box adversarial attacks
  with limited queries and information,'' in \emph{Proceedings of the 35th
  International Conference on Machine Learning, {ICML} 2018,
  Stockholmsm{\"{a}}ssan, Stockholm, Sweden, July 10-15, 2018}, ser.
  Proceedings of Machine Learning Research, J.~G. Dy and A.~Krause, Eds.,
  vol.~80.\hskip 1em plus 0.5em minus 0.4em\relax {PMLR}, 2018, pp. 2142--2151.
  [Online]. Available: \url{http://proceedings.mlr.press/v80/ilyas18a.html}
\BIBentrySTDinterwordspacing

\bibitem{uesato2018adversarial}
\BIBentryALTinterwordspacing
J.~Uesato, B.~O'Donoghue, P.~Kohli, and A.~van~den Oord, ``Adversarial risk and
  the dangers of evaluating against weak attacks,'' in \emph{Proceedings of the
  35th International Conference on Machine Learning, {ICML} 2018,
  Stockholmsm{\"{a}}ssan, Stockholm, Sweden, July 10-15, 2018}, ser.
  Proceedings of Machine Learning Research, J.~G. Dy and A.~Krause, Eds.,
  vol.~80.\hskip 1em plus 0.5em minus 0.4em\relax {PMLR}, 2018, pp. 5032--5041.
  [Online]. Available: \url{http://proceedings.mlr.press/v80/uesato18a.html}
\BIBentrySTDinterwordspacing

\bibitem{cheng2019improving}
S.~Cheng, Y.~Dong, T.~Pang, H.~Su, and J.~Zhu, ``Improving black-box
  adversarial attacks with a transfer-based prior,'' in \emph{Advances in
  Neural Information Processing Systems}, 2019, pp. 10\,934--10\,944.

\bibitem{pmlr-v97-li19g}
Y.~Li, L.~Li, L.~Wang, T.~Zhang, and B.~Gong, ``{NATTACK}: Learning the
  distributions of adversarial examples for an improved black-box attack on
  deep neural networks,'' ser. Proceedings of Machine Learning Research,
  K.~Chaudhuri and R.~Salakhutdinov, Eds., vol.~97.\hskip 1em plus 0.5em minus
  0.4em\relax Long Beach, California, USA: PMLR, 09--15 Jun 2019, pp.
  3866--3876.

\bibitem{su2019one}
J.~Su, D.~V. Vargas, and K.~Sakurai, ``One pixel attack for fooling deep neural
  networks,'' \emph{IEEE Transactions on Evolutionary Computation}, vol.~23,
  no.~5, pp. 828--841, 2019.

\bibitem{Raghunathan18}
\BIBentryALTinterwordspacing
A.~Raghunathan, J.~Steinhardt, and P.~Liang, ``Certified defenses against
  adversarial examples,'' in \emph{6th International Conference on Learning
  Representations, {ICLR} 2018, Vancouver, BC, Canada, April 30 - May 3, 2018,
  Conference Track Proceedings}, 2018. [Online]. Available:
  \url{https://openreview.net/forum?id=Bys4ob-Rb}
\BIBentrySTDinterwordspacing

\bibitem{Dvijotham18}
\BIBentryALTinterwordspacing
K.~Dvijotham, R.~Stanforth, S.~Gowal, T.~A. Mann, and P.~Kohli, ``A dual
  approach to scalable verification of deep networks,'' in \emph{Proceedings of
  the Thirty-Fourth Conference on Uncertainty in Artificial Intelligence, {UAI}
  2018, Monterey, California, USA, August 6-10, 2018}, 2018, pp. 550--559.
  [Online]. Available: \url{http://auai.org/uai2018/proceedings/papers/204.pdf}
\BIBentrySTDinterwordspacing

\bibitem{Wong17}
\BIBentryALTinterwordspacing
E.~Wong and J.~Z. Kolter, ``Provable defenses against adversarial examples via
  the convex outer adversarial polytope,'' in \emph{Proceedings of the 35th
  International Conference on Machine Learning, {ICML} 2018,
  Stockholmsm{\"{a}}ssan, Stockholm, Sweden, July 10-15, 2018}, 2018, pp.
  5283--5292. [Online]. Available:
  \url{http://proceedings.mlr.press/v80/wong18a.html}
\BIBentrySTDinterwordspacing

\bibitem{Salman-NIPS-2019}
H.~Salman, J.~Li, I.~P. Razenshteyn, P.~Zhang, H.~Zhang, S.~Bubeck, and
  G.~Yang, ``Provably robust deep learning via adversarially trained smoothed
  classifiers,'' in \emph{Advances in Neural Information Processing Systems 32:
  Annual Conference on Neural Information Processing Systems 2019, NeurIPS
  2019, 8-14 December 2019, Vancouver, BC, Canada}, H.~M. Wallach,
  H.~Larochelle, A.~Beygelzimer, F.~d'Alch{\'{e}}{-}Buc, E.~B. Fox, and
  R.~Garnett, Eds., 2019, pp. 11\,289--11\,300.

\bibitem{Gowal-Arxiv-2018}
\BIBentryALTinterwordspacing
S.~Gowal, K.~Dvijotham, R.~Stanforth, R.~Bunel, C.~Qin, J.~Uesato,
  R.~Arandjelovic, T.~A. Mann, and P.~Kohli, ``On the effectiveness of interval
  bound propagation for training verifiably robust models,''
  \emph{arXiv:1810.12715}, 2018. [Online]. Available:
  \url{http://arxiv.org/abs/1810.12715}
\BIBentrySTDinterwordspacing

\bibitem{Mirman-ICML-2018}
\BIBentryALTinterwordspacing
M.~Mirman, T.~Gehr, and M.~T. Vechev, ``Differentiable abstract interpretation
  for provably robust neural networks,'' in \emph{Proceedings of the 35th
  International Conference on Machine Learning, {ICML} 2018,
  Stockholmsm{\"{a}}ssan, Stockholm, Sweden, July 10-15, 2018}, ser.
  Proceedings of Machine Learning Research, J.~G. Dy and A.~Krause, Eds.,
  vol.~80.\hskip 1em plus 0.5em minus 0.4em\relax {PMLR}, 2018, pp. 3575--3583.
  [Online]. Available: \url{http://proceedings.mlr.press/v80/mirman18b.html}
\BIBentrySTDinterwordspacing

\bibitem{Wong18}
E.~Wong, F.~Schmidt, J.~H. Metzen, and J.~Z. Kolter, ``Scaling provable
  adversarial defenses,'' in \emph{Advances in Neural Information Processing
  Systems}, 2018, pp. 8400--8409.

\bibitem{Shafahi-NIPS-2019}
\BIBentryALTinterwordspacing
A.~Shafahi, M.~Najibi, A.~Ghiasi, Z.~Xu, J.~P. Dickerson, C.~Studer, L.~S.
  Davis, G.~Taylor, and T.~Goldstein, ``Adversarial training for free!'' in
  \emph{Advances in Neural Information Processing Systems 32: Annual Conference
  on Neural Information Processing Systems 2019, NeurIPS 2019, 8-14 December
  2019, Vancouver, BC, Canada}, H.~M. Wallach, H.~Larochelle, A.~Beygelzimer,
  F.~d'Alch{\'{e}}{-}Buc, E.~B. Fox, and R.~Garnett, Eds., 2019, pp.
  3353--3364. [Online]. Available:
  \url{http://papers.nips.cc/paper/8597-adversarial-training-for-free}
\BIBentrySTDinterwordspacing

\bibitem{Wong-ICLR-2020}
\BIBentryALTinterwordspacing
E.~Wong, L.~Rice, and J.~Z. Kolter, ``Fast is better than free: Revisiting
  adversarial training,'' in \emph{8th International Conference on Learning
  Representations, {ICLR} 2020, Addis Ababa, Ethiopia, April 26-30,
  2020}.\hskip 1em plus 0.5em minus 0.4em\relax OpenReview.net, 2020. [Online].
  Available: \url{https://openreview.net/forum?id=BJx040EFvH}
\BIBentrySTDinterwordspacing

\bibitem{Buckman18}
\BIBentryALTinterwordspacing
J.~Buckman, A.~Roy, C.~Raffel, and I.~Goodfellow, ``Thermometer encoding: One
  hot way to resist adversarial examples,'' in \emph{International Conference
  on Learning Representations}, 2018. [Online]. Available:
  \url{https://openreview.net/forum?id=S18Su--CW}
\BIBentrySTDinterwordspacing

\bibitem{Guo17}
\BIBentryALTinterwordspacing
C.~Guo, M.~Rana, M.~Ciss{\'{e}}, and L.~van~der Maaten, ``Countering
  adversarial images using input transformations,'' in \emph{6th International
  Conference on Learning Representations, {ICLR} 2018, Vancouver, BC, Canada,
  April 30 - May 3, 2018, Conference Track Proceedings}, 2018. [Online].
  Available: \url{https://openreview.net/forum?id=SyJ7ClWCb}
\BIBentrySTDinterwordspacing

\bibitem{Xie17}
\BIBentryALTinterwordspacing
C.~Xie, J.~Wang, Z.~Zhang, Z.~Ren, and A.~L. Yuille, ``Mitigating adversarial
  effects through randomization,'' in \emph{6th International Conference on
  Learning Representations, {ICLR} 2018, Vancouver, BC, Canada, April 30 - May
  3, 2018, Conference Track Proceedings}.\hskip 1em plus 0.5em minus
  0.4em\relax OpenReview.net, 2018. [Online]. Available:
  \url{https://openreview.net/forum?id=Sk9yuql0Z}
\BIBentrySTDinterwordspacing

\bibitem{Song17}
\BIBentryALTinterwordspacing
Y.~Song, T.~Kim, S.~Nowozin, S.~Ermon, and N.~Kushman, ``Pixeldefend:
  Leveraging generative models to understand and defend against adversarial
  examples,'' in \emph{6th International Conference on Learning
  Representations, {ICLR} 2018, Vancouver, BC, Canada, April 30 - May 3, 2018,
  Conference Track Proceedings}, 2018. [Online]. Available:
  \url{https://openreview.net/forum?id=rJUYGxbCW}
\BIBentrySTDinterwordspacing

\bibitem{Samangouei18}
\BIBentryALTinterwordspacing
P.~Samangouei, M.~Kabkab, and R.~Chellappa, ``Defense-gan: Protecting
  classifiers against adversarial attacks using generative models,'' in
  \emph{6th International Conference on Learning Representations, {ICLR} 2018,
  Vancouver, BC, Canada, April 30 - May 3, 2018, Conference Track Proceedings},
  2018. [Online]. Available: \url{https://openreview.net/forum?id=BkJ3ibb0-}
\BIBentrySTDinterwordspacing

\bibitem{Raff19}
E.~Raff, J.~Sylvester, S.~Forsyth, and M.~McLean, ``Barrage of random
  transforms for adversarially robust defense,'' in \emph{Proceedings of the
  IEEE Conference on Computer Vision and Pattern Recognition}, 2019, pp.
  6528--6537.

\bibitem{Maung-Access-2019}
M.~{AprilPyone}, Y.~{Kinoshita}, and H.~{Kiya}, ``Adversarial robustness by one
  bit double quantization for visual classification,'' \emph{IEEE Access},
  vol.~7, pp. 177\,932--177\,943, 2019.

\bibitem{Metzen-ICLR-2017}
J.~H. Metzen, T.~Genewein, V.~Fischer, and B.~Bischoff, ``On detecting
  adversarial perturbations,'' in \emph{5th International Conference on
  Learning Representations, {ICLR} 2017, Toulon, France, April 24-26, 2017,
  Conference Track Proceedings}, 2017.

\bibitem{Feinman-Arxiv-2017}
\BIBentryALTinterwordspacing
R.~Feinman, R.~R. Curtin, S.~Shintre, and A.~B. Gardner, ``Detecting
  adversarial samples from artifacts,'' \emph{arXiv:1703.00410}, 2017.
  [Online]. Available: \url{http://arxiv.org/abs/1703.00410}
\BIBentrySTDinterwordspacing

\bibitem{Carlini-AISec-2017}
\BIBentryALTinterwordspacing
N.~Carlini and D.~A. Wagner, ``Adversarial examples are not easily detected:
  Bypassing ten detection methods,'' in \emph{Proceedings of the 10th {ACM}
  Workshop on Artificial Intelligence and Security, AISec@CCS 2017, Dallas, TX,
  USA, November 3, 2017}, B.~M. Thuraisingham, B.~Biggio, D.~M. Freeman,
  B.~Miller, and A.~Sinha, Eds.\hskip 1em plus 0.5em minus 0.4em\relax {ACM},
  2017, pp. 3--14. [Online]. Available:
  \url{https://doi.org/10.1145/3128572.3140444}
\BIBentrySTDinterwordspacing

\bibitem{Taran-ECCV-2018}
O.~Taran, S.~Rezaeifar, and S.~Voloshynovskiy, ``Bridging machine learning and
  cryptography in defence against adversarial attacks,'' in \emph{Proceedings
  of the European Conference on Computer Vision (ECCV)}, 2018.

\bibitem{LeCun-IEEE-1998}
Y.~LeCun, L.~Bottou, Y.~Bengio, and P.~Haffner, ``Gradient-based learning
  applied to document recognition,'' \emph{Proceedings of the IEEE}, vol.~86,
  no.~11, pp. 2278--2324, 1998.

\bibitem{Xiao-preprint-2017}
\BIBentryALTinterwordspacing
H.~Xiao, K.~Rasul, and R.~Vollgraf, ``Fashion-mnist: a novel image dataset for
  benchmarking machine learning algorithms,'' \emph{arXiv:1708.07747}, 2017.
  [Online]. Available: \url{http://arxiv.org/abs/1708.07747}
\BIBentrySTDinterwordspacing

\bibitem{Krizhevsky09}
A.~Krizhevsky and G.~Hinton, ``Learning multiple layers of features from tiny
  images,'' University of Toronto, Tech. Rep., 2009.

\bibitem{ILSVRC15}
O.~Russakovsky, J.~Deng, H.~Su, J.~Krause, S.~Satheesh, S.~Ma, Z.~Huang,
  A.~Karpathy, A.~Khosla, M.~Bernstein, A.~C. Berg, and L.~Fei-Fei, ``{ImageNet
  Large Scale Visual Recognition Challenge},'' \emph{International Journal of
  Computer Vision (IJCV)}, vol. 115, no.~3, pp. 211--252, 2015.

\bibitem{Bellare-NIST-2010}
M.~Bellare, P.~Rogaway, and T.~Spies, ``Addendum to “the ffx mode of
  operation for format-preserving encryption”,'' \emph{A parameter collection
  for enciphering strings of arbitrary radix and length, Draft 1.0, NIST},
  2010.

\bibitem{He16}
K.~He, X.~Zhang, S.~Ren, and J.~Sun, ``Deep residual learning for image
  recognition,'' in \emph{Proceedings of the IEEE conference on computer vision
  and pattern recognition}, 2016, pp. 770--778.

\bibitem{Smith-Arxiv-2017}
\BIBentryALTinterwordspacing
L.~N. Smith and N.~Topin, ``Super-convergence: Very fast training of residual
  networks using large learning rates,'' \emph{arXiv:1708.07120}, 2017.
  [Online]. Available: \url{http://arxiv.org/abs/1708.07120}
\BIBentrySTDinterwordspacing

\bibitem{Micikevicius-Arxiv-2017}
\BIBentryALTinterwordspacing
P.~Micikevicius, S.~Narang, J.~Alben, G.~F. Diamos, E.~Elsen, D.~Garc{\'{\i}}a,
  B.~Ginsburg, M.~Houston, O.~Kuchaiev, G.~Venkatesh, and H.~Wu, ``Mixed
  precision training,'' \emph{arXiv:1710.03740}, 2017. [Online]. Available:
  \url{http://arxiv.org/abs/1710.03740}
\BIBentrySTDinterwordspacing

\bibitem{ding2019advertorch}
\BIBentryALTinterwordspacing
G.~W. Ding, L.~Wang, and X.~Jin, ``advertorch v0.1: An adversarial robustness
  toolbox based on pytorch,'' \emph{arXiv:1902.07623}, 2019. [Online].
  Available: \url{http://arxiv.org/abs/1902.07623}
\BIBentrySTDinterwordspacing

\bibitem{2019-NIPS-Zhang}
H.~Zhang and J.~Wang, ``Defense against adversarial attacks using feature
  scattering-based adversarial training,'' in \emph{Advances in Neural
  Information Processing Systems}, 2019.

\end{thebibliography}

\end{document}